\documentclass[11pt]{article}
\usepackage[top=1 in, bottom=1 in, left=0.85 in, right=0.85 in]{geometry}

\usepackage{amsmath,amsfonts,bm}

\def\eqref#1{equation~\ref{#1}}

\def\1{\bm{1}}

\def\vmu{{\bm{\mu}}}
\def\vtheta{{\bm{\theta}}}

\def\ve{{\bm{e}}}

\def\vg{{\bm{g}}}

\def\vv{{\bm{v}}}

\def\vx{{\bm{x}}}
\def\vy{{\bm{y}}}

\def\mP{{\bm{P}}}

\DeclareMathAlphabet{\mathsfit}{\encodingdefault}{\sfdefault}{m}{sl}
\SetMathAlphabet{\mathsfit}{bold}{\encodingdefault}{\sfdefault}{bx}{n}

\def\gD{{\mathcal{D}}}
\def\gE{{\mathcal{E}}}
\def\gF{{\mathcal{F}}}

\def\gI{{\mathcal{I}}}

\def\gN{{\mathcal{N}}}
\def\gO{{\mathcal{O}}}
\def\gP{{\mathcal{P}}}

\def\sA{{\mathbb{A}}}

\def\sP{{\mathbb{P}}}

\def\sR{{\mathbb{R}}}
\def\sS{{\mathbb{S}}}

\newcommand{\E}{\mathbb{E}}

\DeclareMathOperator{\sign}{sign}

\usepackage{float}
\usepackage{times}
\usepackage{mathtools}
\usepackage{makecell}
\usepackage{amsmath,amssymb,graphicx,url}
\usepackage{thmtools,thm-restate,wrapfig,enumitem,mathabx}
\usepackage{booktabs}
\usepackage{color}
\usepackage{colortbl}
\usepackage{multirow}

\usepackage[utf8]{inputenc}
\usepackage[T1]{fontenc}
\usepackage{hyperref}
\usepackage{url}
\usepackage{booktabs}
\usepackage{amsfonts}
\usepackage{nicefrac}
\usepackage{microtype}
\usepackage{xcolor}

\usepackage{caption}
\usepackage{subcaption}

\usepackage[ruled,vlined,linesnumbered]{algorithm2e}
\SetKwBlock{Init}{Initialization: }{}
\usepackage{titletoc}
\usepackage{bbm}
\usepackage{diagbox}
\usepackage{tikz}
\usetikzlibrary{calc}

\newtheorem{assumption}{Assumption}
\newtheorem{theorem}{Theorem}

\newtheorem{lemma}{Lemma}
\newtheorem{corollary}{Corollary}

\newtheorem{definition}{Definition}
\newtheorem{proposition}{Proposition}

\title{Efficient Federated RLHF via Zeroth-Order Policy Optimization}
\author{
    Deyi Wang\thanks{Both authors contributed equally to this research.} \\
	University of Michigan, Ann Arbor\\
	\texttt{deyiwang@umich.edu} \\
	\and
	Qining Zhang\footnotemark[1] \\
	University of Michigan, Ann Arbor\\
	\texttt{qiningz@umich.edu} \\
    \and
	Lei Ying \\
	University of Michigan, Ann Arbor\\
	\texttt{leiying@umich.edu} \\
}
\date{}
\begin{document}

\maketitle

\begin{abstract}
This paper considers reinforcement learning from human feedback in a federated learning setting with resource-constrained agents, such as edge devices. We propose an efficient federated RLHF algorithm, named Partitioned, Sign-based Stochastic Zeroth-order Policy Optimization (\texttt{Par-S$^2$ZPO}). The algorithm is built on zeroth-order optimization with binary perturbation, resulting in low communication, computation, and memory complexity by design. Our theoretical analysis establishes an upper bound on the convergence rate of \texttt{Par-S$^2$ZPO}, revealing that it is as efficient as its centralized counterpart in terms of sample complexity but converges faster in terms of policy update iterations. Our experimental results show that it outperforms a FedAvg-based RLHF on four MuJoCo RL tasks. 
\end{abstract}

\section{Introduction}

Reinforcement Learning from Human Feedback (RLHF) has become a cornerstone for fine-tuning LLMs~\cite{OuyWuJia_22,RafShaMit_23} and advancing physical intelligence in robotics~\cite{ChrLeiBro_17,kim2025}. This paper considers RLHF in a federated learning setting, in which multiple resource-constrained agents (e.g., edge devices) collaborate to train an RL model using human preferences. Implementing existing federated algorithms, such as FedAvg~\cite{McmMooRam_17}, for this purpose requires frequent computation and exchanges of a large amount of gradient information, and is therefore challenging when edge devices have limited memory, computation, and communication resources. For example, fine-tuning an LLM involves the gradients of billions of model parameters, which could exceed the communication and computing capacity of an edge device. Even when such fine-tuning is feasible, it may also quickly drain the battery. 

In this paper, we propose a federated RLHF algorithm, named {\em Partitioned, Sign-based Stochastic Zeroth-order Policy Optimization}, or \texttt{Par-S$^2$ZPO} for short.
It is efficient in communication, computation, and memory by design, based on the following key ideas:
\begin{itemize}[leftmargin=*]
    \item \textbf{Zeroth-order optimization with binary perturbation:} \texttt{Par-\allowbreak S$^2$ZPO} uses the following sign-based, two-point, zeroth-order method to approximate the direction of the gradient: $$\nabla_\vtheta V(\pi_\vtheta)\approx  c_0~\mathbb{E}_\vv\left[\sign[V(\pi_{\vtheta+\mu \vv})-V(\pi_\vtheta)] \vv\right],$$ where $\pi_{\vtheta}$ is the current policy parameterized by a parameter $\vtheta$, $\pi_{\vtheta+\mu \vv}$ is a perturbed policy obtained by perturbing the policy parameter with a direction $\vv$ and distance $\mu$, and $c_0$ is an absolute constant. To reduce communication and memory complexity, \texttt{Par-S$^2$ZPO} uses binary perturbations, in which each entry is either $+1$ or $-1$, so $v$ can be {\em stored and communicated easily}.

    \item \textbf{Partitioned RLHF:} To further reduce the memory requirement at each agent, \texttt{Par-S$^2$ZPO} partitions the policy parameters into $K$ subsets, where $K$ is the number of agents. Each agent perturbs only the parameter subset assigned to them: that is, for agent $k,$ the perturbed policy has parameter ${\vtheta}+\mu \vv_k,$ where the $i$-th element of $
    \vv_k$ is the $i$-th element of $\vv$ only if the this parameter entry is assigned to agent $k$. Otherwise, it is $0$. For example, if the actor is a neural network with $mK$ layers, then \texttt{Par-S$^2$ZPO} assigns $m$ layers to each agent.  

    \item \textbf{Binary feedback: } After agent $k$ compares the current policy with its perturbed policy from human feedback, the agent communicates only a binary value $\hat{O}_k \in \{-1,+1\}$ indicating whether $\vv_k$ is a favorable perturbation direction, incuring minimal communication overhead. The policy is moved toward direction $\hat{\vg}=\sum_k \hat{O}_k\vv_k$ with a properly chosen learning rate $\alpha$. 
\end{itemize}

\subsection{Main Contributions}
The key contributions of this paper are summarized below. 

\paragraph{Efficient Algorithm Design.}Assume the parameterized policy consists of $d$ parameters, i.e., $\vtheta \in \sR^d,$. During each policy update, \texttt{Par-S$^2$ZPO} requires only $d+K$ bits of communication, each agent needs to store only an additional $d/K$ parameters besides the current policy, and furthermore, no gradient calculation is needed. Therefore, \texttt{Par-S$^2$ZPO} is communication-, computation-, and memory-efficient.

\paragraph{Provable Rate of Convergence.} To understand its performance, we theoretically analyze the convergence of \texttt{Par-S$^2$ZPO} and establish an upper bound on the convergence rate. From the upper bound, we have two key observations:
\begin{itemize}
    \item When fixing the total number of trajectories for human evaluations during training, which can be regarded as the sample complexity of the algorithm, the upper bound is independent of $K.$ In other words, sample-wise, a $K$-agent system is as efficient as a centralized, single-agent system ($K=1$), even when the single-agent system takes $K$ times more iterations under the same sample complexity. Our simulations will confirm this observation.

    \item There is a tradeoff in choosing the batch size for obtaining human preference feedback because a large batch size results in more accurate evaluations from human evaluators, but a smaller number of training iterations. In practice, this is an important hyperparameter that should be chosen carefully based on the environment.
\end{itemize}

\paragraph{Numerical Evaluation.} Empirically, we implemented \texttt{Par-S$^2$ZPO} in four MuJoCo RL environments, evaluated its performance with different $K$s, and compared it with FedAvg-based zeroth-order policy optimization. \texttt{Par-S$^2$ZPO} outperforms the FedAvg-based algorithm in all four environments. 

\subsection{Related Work}

The proposed \texttt{Par-S$^2$ZPO} is related to several research areas.

\textbf{RLHF and preference-based RL:} RLHF methods, such as InstructGPT~\cite{OuyWuJia_22} and DPO~\cite{RafShaMit_23}, have become a cornerstone of post-training for large language models (LLMs). Recently, zeroth-order methods such as ZPG~\cite{ZhaYin_25} and ZSPO~\cite{ZhaYin_25_Unknown} have also been proposed for RLHF in general MDPs. ZPG~\cite{ZhaYin_25} estimates value differences from human preferences and combines them with zeroth-order gradient approximation, whereas ZSPO~\cite{ZhaYin_25_Unknown} only estimates the sign of the value difference via human preferences, so it works for unknown preference models (link functions). \texttt{Par-S$^2$ZPO} extends zeroth-order policy optimization to the federated setting. Unlike ZSPO and ZPG, which use real-valued perturbation vectors (e.g., Gaussian or uniform), \texttt{Par-S$^2$ZPO} uses Rademacher perturbations ($\{-1,+1\}$) to reduce communication, computation, and memory costs. As a result, its convergence analysis differs.

\textbf{Federated Learning (FL) and Split Learning (SL):} FL~\cite{McmMooRam_17} enables multiple agents to collaboratively train a model while keeping data local. In FedAvg~\cite{McmMooRam_17}, the central server updates the global model by aggregating local updates from all agents. FedBCD~\cite{fedbcd} adopts a block-coordinate training strategy, in which each agent updates only a subset of the model parameters. Split Learning (SL)~\cite{vepakomma2018split} goes one step further by partitioning a model into multiple components and storing only one component at each agent, and requires sequential communication across the split model during training. Similar to FedBCD~\cite{fedbcd}, \texttt{Par-S$^2$ZPO} partitions the parameter set into subsets. Unlike FedBCD~\cite{fedbcd} and SL \cite{vepakomma2018split}, \texttt{Par-S$^2$ZPO} uses zeroth-order policy optimization with binary perturbations rather than gradient-based updates, thereby reducing computation and communication overhead. 

\textbf{Sign-based Zeroth-Order Optimization:} Sign-based methods, such as sign-GD~\cite{LiuCheChe_19}, estimate coordinate-wise gradient signs and use them to optimize the objective. In contrast, \texttt{Par-S$^2$ZPO} uses binary perturbations together with preference feedback to infer a policy-improvement direction from human preferences. Therefore, the ``sign'' used in \texttt{Par-S$^2$ZPO} is conceptually different from sign-GD.

\section{Preliminaries}

In this section, we introduce the problem setting, the partitioned federated learning scheme, and the necessary notations.

\subsection{Preference-based Reinforcement Learning}

We first present the model of the episodic RL problem, and then introduce the preference feedback mechanism.

\textbf{Episodic RL.} An episodic RL problem is represented by a finite-horizon Markov decision process $\mathcal{M} = (\sS, \sA, H, \mP, \vmu_0)$, where $\sS$ is the state space and $\sA$ is the action space, both may be uncountable. $H$ is the planning horizon, $\mP = \{\mP_h\}_{h=1}^H$ is the set of transition kernels, and $\vmu_0$ is the initial distribution of states. 
At the start of each episode $t$, the agent chooses a policy $\pi_{t}$ which is represented as a set of functions $\{\pi_{t,h}:\sS \to \mathcal{P}(\sA)\}_{h=1}^H$, where $\mathcal{P}(\sA)$ denotes the set of all probability distributions over the action space. Then an initial state $s_{t,1}$ is sampled from the initial distribution $\vmu_0$. At each step $h$, the agent observes the current state $s_{t,h}$ and then takes an action $a_{t,h} = \pi_{h}(s_{t,h})$ according to the chosen policy. The environment consequently moves to a new state $s_{t,h+1}$ sampled from the distribution $\mP_h(\cdot|s_{t,h}, a_{t,h}).$ 
For each trajectory $\tau_{t} = \{(s_{t,h}, a_{t,h})\}_{h=1}^H$ generated by the agent at episode $t$, we assume the expected return $r(\tau_{t})\in[0, H]$ is a general function determined by the entire trajectory and may not be possible to be decomposable into per-step rewards as in traditional RL. For any policy $\pi$, we define the value function as $J(\pi) = \E_{s\sim \vmu_0}[V^\pi(s)]$ where $V^{\pi}(s)$ is the expected reward of trajectories starting from $s$ and using policy $\pi$:
\begin{align*}
    V^\pi(s) =& \E_\pi[ \left. r(\tau)\right| s_{1} = s] = \E[ \left. r(\tau)\right| s_{1} = s, \{a_{1},\cdots,a_{H}\} \sim \pi ].
\end{align*}

\textbf{Policy Parameterization and Optimization.} We consider parametrized policies, represented as a function class $\gN = \{\pi_{\vtheta}: \sS \times [H] \to \mathcal{P}(\sA) | \vtheta \in \sR^d\}$, which takes a step $h$ and state $s$ as input, and outputs the probability distribution of the next action. For example, $\vtheta$ may be the weights of an actor network. Each parameter $\vtheta$ induces a policy denoted as $\pi_\vtheta$.
Our goal is to find the best parameter $\vtheta^*$ to maximize the expected value function over the initial state distribution, i.e.,
$$
    \max_{\vtheta \in \sR^d} V(\pi_{\vtheta}) = \E_{s\sim \vmu_0}[V^{\pi_\vtheta}(s)].
$$

\textbf{Preference Feedback.} The agent has access to a panel of $P$ homogeneous preference panelists. The agent can choose two batches of trajectories $\gD_{0} = \{\tau_{0,i}\}_{i=0}^D$ and $\gD_{1} = \{\tau_{1,i}\}_{i=1}^D$, each of which containing $D$ trajectories, to query preferences (one-bit feedback $o^p \in\{0,1\}$), from each panelist $p=1,2,\dots.P$, where $0$ means $\gD_0$ is preferred and $1$ means $\gD_1$ is preferred. We assume a utility-based generation mechanism. Specifically, a panelist generates the preference according to a preference model characterized by a \emph{link} function $\sigma(\cdot)$ of the average reward difference between trajectories:
\begin{align*}
    \sP(o^p=1) = \sigma\left(\frac{1}{D}\sum_{i=1}^D r(\tau_{i,1}) - \frac{1}{D}\sum_{i=1}^D r(\tau_{i,0})\right).
\end{align*}
Most preference-based RL works assume either the Bradley-Terry model, in which $\sigma(\cdot)$ is a standard logistic function~\cite{ChrLeiBro_17}, or a linear model, in which $\sigma(\cdot)$ is a linear function~\cite{BenBusMes_21}. We assume a general and unknown link function and the agent receives a one-bit preference feedback~\cite{ZhaYin_25_Unknown}. Note that a panel contains $P$ panelists, and the panel's preference $o$ follows the majority vote among panelists: $o=\mathbbm{1}\{\sum_{p=1}^Po^p>\frac{P}{2}\},$ where $\mathbbm{1}\{\cdot\}$ is the indicator function. The panel preference feedback $o$ can also be characterized by a link function denoted by $\sigma_P$, $$\sP(o=1)=\sP\left(\sum_{p=1}^Po^p>\frac{P}{2}\right)=\sigma_P\left(\frac{1}{D}\sum_{i=1}^D r(\tau_{i,1}) - \frac{1}{D}\sum_{i=1}^D r(\tau_{i,0})\right).$$ $\sigma_P$ is uniquely determined by $\sigma.$ Larger $P$, having more panelists in a panel, increases the accuracy of the preference, and the link function $\sigma_P$ will be closer to a step function.

\subsection{Notations}

Let $\ve_i\in \sR^d$ represent the unit vector with all zero elements but $1$ on the $i$-th coordinate. For a scalar $a$,  we use $\sign[a]$ to denote the sign of the scalar, i.e., $\sign[a] = 1$ if $a> 0$, $\sign[a] = -1$ if $a<0$, and $\sign[0]=0$. For a vector $\vv$, $\sign[\vv]$ represents the vector after applying the $\sign$ operator entry-wise. We use $\top$ to denote the transpose. To analyze the partial parameters, we introduce the mask vector and the Hadamard product operator. Given a coordinate indices subset $\gI\subseteq\{1,2,\dots,d\}$, we define the binary mask vector $\bm{e}_{\gI}\in \{0,1\}^d$ whose $i$-th entry is $1$ if $i \in \gI$, and $0$ otherwise. The Hadamard product $\bm{a}\circ\bm{b}$ of two vectors $\bm{a}$ and $\bm{b}$ is the entry-wise product of $\bm{a}$ and $\bm{b}$. Hence, for any given vector $\bm{a}\in\sR^d$, $\bm{a}\circ\bm{e}_{\gI}$ is the masked vector with the entries of $\bm{a}$ not in ${\gI}$ being $0$.

\subsection{Partitioned Federated Reinforcement Learning (ParFed-RL)}

\begin{figure}[h]
    \centering
    \includegraphics[width=0.5\linewidth]{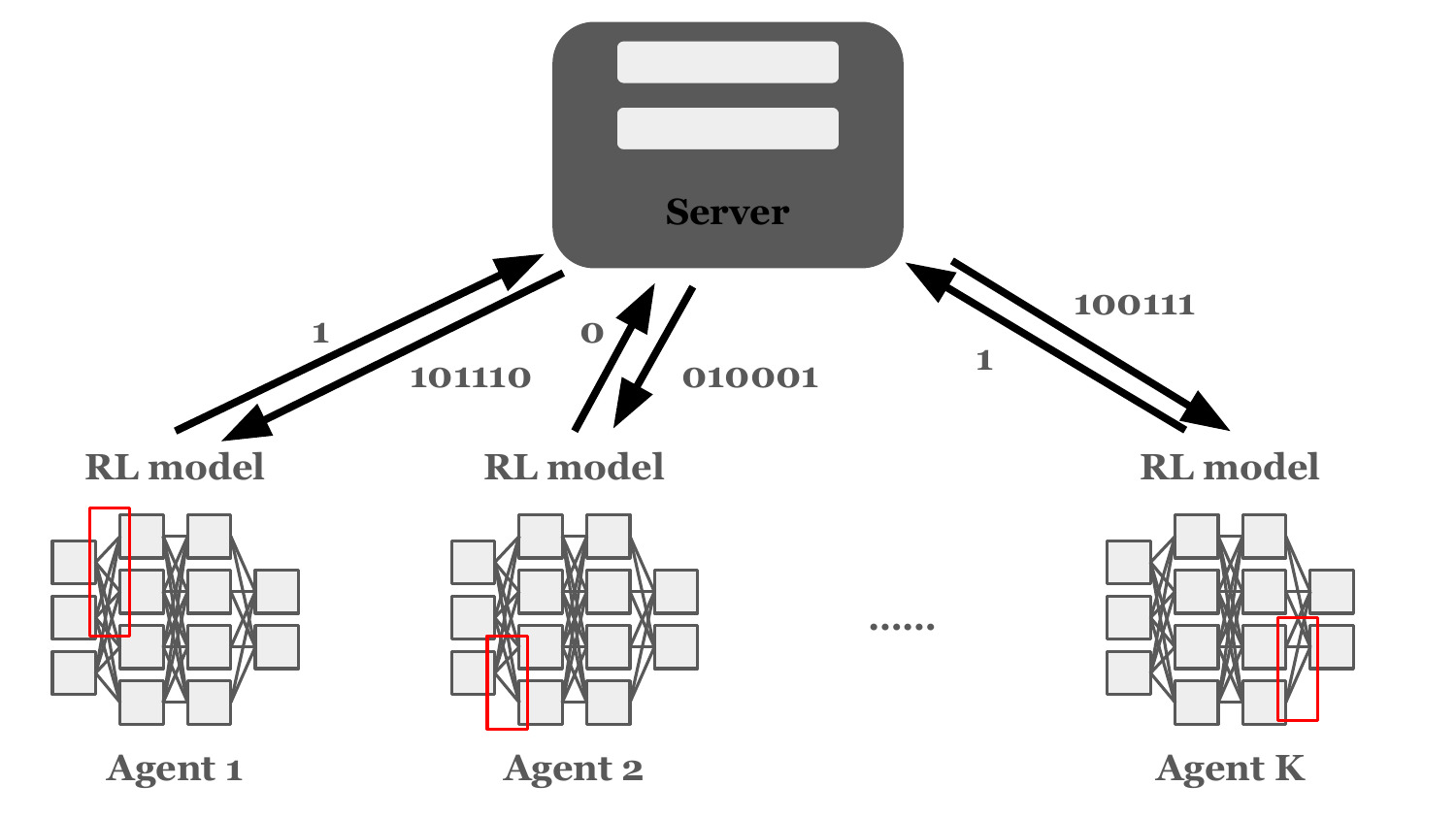}
    \caption{ParFed RL}
    \label{fig:frl}
\end{figure}

In this section, we introduce the partitioned federated reinforcement learning (ParFed-RL) framework as shown in Figure \ref{fig:frl}, which consists of a central server and $K$ agents. Each agent interacts with their respective local environment to collect trajectories. Then, it queries the local preference panelists for preference feedback over pairs of trajectory batches. Note that in the existing federated reinforcement learning algorithms~\cite{fan2024}, each agent computes an update direction and communicates it to the server, and the server aggregates the updates, e.g., by averaging, and updates the model. In ParFed-RL, each agent is responsible for updating just a subset of parameters instead of computing an update direction for the entire parameter vector.  In particular, assuming the parameter $\vtheta$ is $d$-dimensional, ParFed-RL partitions it into $K$ equal-sized blocks, $\vtheta=[\vtheta^{(1)},\cdots,\vtheta^{(K)}]^\top$, where each $\vtheta^{(k)} \in \sR^{d/K}$ is a block of parameters.
Agent $k$ then computes a partial ascent direction ${\vv}_k\circ{\ve}_{\gI_k},$ where $\gI_k$ is the index set of the assigned parameter block to agent $k,$ and broadcasts it. The server and all agents then update their actors with learning rate $\alpha:$
$$\bm{\theta}\leftarrow \bm{\theta}+\alpha\sum_k {\vv}_k\circ{\ve}_{\gI_k}.$$ 
If we let ${\vv}_k\circ{\ve}_{\gI_k}$ be the partial gradient, then this becomes a multi-agent version of the policy-gradient algorithm. However, computing the gradient is both computationally and memory-intensive, and communicating the exact gradients also requires significant bandwidth. We next present a ParFed-RL algorithm that is based on stochastic zeroth-order policy optimization with binary perturbation and has low communication overhead and low computation and memory complexity.

\section{Algorithm}

In this section, we introduce \texttt{Par-S$^2$ZPO}, which stands for {\it Partitioned, Sign-based Stochastic Zeroth-order Policy Optimization}.
The algorithm is shown in  Algorithm \ref{algorithm1}. 

\begin{algorithm}
    \caption{Partitioned, Sign-based Stochastic Zeroth-order Policy Optimization (\texttt{Par-S$^2$ZPO})}
    \label{algorithm1}
    \Init{
    \textbf{Central server: }initialize the actor-network parameter ${\vtheta}_1 \in \sR^d$, create parameter coordianate partition $\gI_1$, $\gI_2$, $\cdots$, $\gI_K$, and assign each agent one subset. Initialize the learning rate scheduler $\{\alpha_t\}_{t=1}^T$ and perturbation distances $\{\mu_t\}_{t=1}^T$, and broadcast to each agent.
    }
    
    \For{policy iteration $t = 1$ \KwTo $T$}{
        \tcp*[h]{\textbf{Sample perturbation direction}}\;
        sample a $d$-dimensional random vector ${\vv}_t$ from an independent Rademacher distribution\;
        broadcast $ \vv_t$ to all agents.\; 
        \tcp*[h]{\textbf{Partitioned policy evaluation}}\;
        \For{agent $k=1$ \KwTo $K$}{
            compute ${\vv}_{t,k}={\vv}_t\circ{\ve}_{\gI_k}$ from $ \vv_t$\;
            set perturbed parameter ${\vtheta}'_{t,k} = {\vtheta}_t + \mu_{t} {\vv}_{t,k}$\;
            evaluate both the policy ${\vtheta}_t$ and the perturbed policy ${\vtheta}'_{t,k}$ with the local preference panel with Algorithm \ref{oracle}, and obtain the majority-voted feedback $\hat{{O}}_{t,k}$\;
            broadcast $\hat{{O}}_{t,k}$ to the central server and other agents\;
        }
        \tcp*[h]{\textbf{Policy update}}\;
        The server and all agents aggregate partial update directions from all agents $\hat{{\vg}}_t=\sum_{k=1}^K\hat{{O}}_{t,k}{\vv}_{t,k}$ and 
update their actor networks ${\vtheta}_{t+1}={\vtheta}_t+\alpha_t\hat{{\vg}}_t$.
}
\end{algorithm}

At each iteration $t$, \texttt{Par-S$^2$ZPO} consists of the following key steps:
\begin{enumerate}[leftmargin=*]
    \item \textbf{Perturbation Sampling.} The central server generates a $d$-dimensional parameter perturbation vector $ \vv_t$, where each element is independently sampled from a symmetric binary distribution over $\{-1,1\}$, i.e., a Rademacher distribution, which will be broadcast to all agents.
    \item \textbf{Policy Perturbation.} After receiving $ \vv_t,$ agent $k$ perturbs the subset of parameters it is responsible for based on $ \vv_t$ to obtain a perturbed policy actor $\vtheta_{t,k}$. 
    \item \textbf{Policy Evaluation. } For both policies $\vtheta_t$ and $\vtheta_{t,k}$, agent $k$ interacts with the environment to sample multiple batches of trajectories and queries the local preference panel to obtain the preference over policies ($\hat{{O}}_{t,k}\in\{-1,1\}$) through a majority vote. The mechanism is shown in Algorithm \ref{oracle}.
    \item \textbf{Preference Aggregation and Policy Update.} After each agent broadcasts the policy preference $\hat{{O}}_{t,k}$, the central server and all agents compute the binary policy parameter update direction by stacking the signed perturbation vector, i.e.,
    \begin{align*}
        \hat{\vg}_t = \sum_{k=1}^K \hat{{O}}_{t,k} \left( \vv_t \circ{\ve}_{\gI_k} \right),
    \end{align*} and obtain a new actor via update $\vtheta_{t+1}=\vtheta_t+\alpha_t\hat{\vg}_t.$ 
\end{enumerate}

\begin{algorithm}
    \caption{Mechanism of local preference panel}
    \label{oracle}
    \KwIn{A policy $\pi_{{\vtheta}}$ and a perturbed policy $\pi_{{\vtheta}'}$}
    \For{$n = 1$ \KwTo $N$}{
        sample a batch of $D$ trajectories $\mathcal{D}_{n,0} \sim \pi_{{\vtheta}}$\;
        sample a batch of $D$ trajectories $\mathcal{D}_{n,1} \sim \pi_{{\vtheta}'}$\;
        \textcolor{black}{query the preference panel with $P$ oracles over $(\mathcal{D}_{n,1}, \mathcal{D}_{n,0})$ and obtain result $o^p_{n}\in\{0,1\}$ for $p=1,2,\dots,P$, and then obtain the panel preference feedback $o_n\in\{0,1\}$ via in-panel majority vote}\;
    }
    estimate the gradient ascent direction with majority vote: $\hat{{O}} = \operatorname{sign} \left[ \left( \sum_{n=1}^{N} \left( o_{n} - \frac{1}{2} \right) \right)\right]$\;
    \KwOut{The majority vote $\hat{{O}}.$}
\end{algorithm}

The design of \texttt{Par-S$^2$ZPO} has two ingredients: partition of the parameters, zeroth-order optimization with binary perturbation.  

\textbf{Zeroth-order optimization with binary perturbation:} To avoid communicating and storing real-valued gradients, \texttt{Par-S$^2$ZPO} is based on zeroth-order optimization \cite{NesSpo_17}, where the key idea is estimating the gradient by the value comparison between two points by perturbing $\vx$ to $\vx+\mu \vv$: $$\nabla f(\vx)\approx  c_0~\mathbb{E}_{\vv}\left[\sign[f(\vx+\mu \vv)-f(\vx)] \vv\right].$$  \texttt{Par-S$^2$ZPO} uses {\em binary} perturbation instead of Gaussian perturbation, so it is communication efficient. 
Since it is a gradient-free property, it further reduces computation and memory complexity. 

\textbf{Parameter Partition:} The federated learning framework enables the collaboration among distributed agents in training one RL model together. To further leverage the diversity of different agents, \texttt{Par-S$^2$ZPO} partitions the parameter set into $K$ subsets and each agent focuses on evaluating the advantage of the subset perturbation, which gives $2^K$ possible directions, i.e., $(\pm \vv_{t,1}, \pm \vv_{t,2}, \cdots, \pm \vv_{t, K})$ instead of $2$ directions, $\pm  \vv_t$ in the centralized case.  

\section{Main Results}

In this section, we quantify the convergence rate of \texttt{Par-S$^2$ZPO}. 
\subsection{Definitions and Assumptions}

To measure the performance of \texttt{Par-S$^2$ZPO}, which updates the actor network block-wise, we first define the block-sum norm.

\begin{definition}\label{fnorm}
    \textbf{(Block-sum norm)} For any $d$-dimension vector $\vv\in\sR^d$, given a partition of the coordinate indices $\gP_K=\{\gI_k\}_{k=1}^K$ with $K$ subsets, we define the block-sum norm as:
    $$\left\|\vv\right\|_{\gP_K}:=\sum_{k=1}^K\sqrt{\sum_{i\in\gI_k}v_i^2},$$ where $v_i$ is the $i$-th entry of $\vv$.
\end{definition}

\begin{proposition}\label{pro1}
    The block-sum norm $\left\|\cdot\right\|_{\gP_K}$ is a norm in $\sR^d$, i.e.,
    \begin{enumerate}
    \item for any $\vx,\vy$, $\left\|\vx+\vy\right\|_{\gP_K}\leq\left\|\vx\right\|_{\gP_K}+\left\|\vy\right\|_{\gP_K}.$
    \item for any $\vx$ and scalar $a$, $\left\|a\vx\right\|_{\gP_K}=|a|\cdot\left\|\vx\right\|_{\gP_K}.$
    \item for $\vx$, if $\left\|\vx\right\|_{\gP_K}=0$, then $\vx=\bm{0}$.
\end{enumerate}
\end{proposition}

\begin{proposition}\label{pro2}
    Given a partition of the coordinate indices $\gP_K$, $$\left\|\vv\right\|_2\leq\left\|\vv\right\|_{\gP_K}\leq\sqrt{K}\left\|\vv\right\|_2, \quad \forall \vv \in \sR^d.$$
\end{proposition}
These two propositions can be easily proved via the Cauchy-Schwarz inequality. We then assume the value function in our problem is smooth, which is a popular assumption used in analyzing deep learning algorithms~\cite{Bottou2018,Jinghui2018,Sanjeev2018} and RL algorithms~\cite{Pirotta2015,Lingxiao2019}.

\begin{assumption}\label{smoothvaluefunction}
    The value function $V\left(\pi_\vtheta\right)$ is $L$-smooth, i.e., it satisfies  $$\left\|\nabla_\theta V\left(\pi_{\vtheta_1}\right)-\nabla_\theta V\left(\pi_{\vtheta_2}\right)\right\|_2\leq L\left\|\vtheta_1-\vtheta_2\right\|_2$$
\end{assumption}

The smoothness assumption implies that a small perturbation in the parameters results in only a small change in the value function. Under this assumption, the panelists may find it difficult to distinguish between two policies by just comparing a finite batch of trajectories when their values are close to each other.
To model this,  we introduce distinguishability, which characterizes the limitation of the human preference when evaluating finite batches~\cite{ZhaYin_25_Unknown}.

\begin{definition}\label{HFlimitation}
    \textbf{(Distinguishability)}  Let $D$ be the batch size, and $\varsigma_P\left(x\right)=\sigma_P\left(x\right)-\frac{1}{2}$ be the preference deviation function. The distinguishability, $\epsilon_D^*>0,$ is the maximum constant $\epsilon$ such that for any two policies $\pi_1$ and $\pi_2$ with $V\left(\pi_1\right)-V\left(\pi_2\right)>\epsilon$, $$\E_{\gD_1\sim\pi_1,\gD_2\sim\pi_2}\left[\varsigma_P\left(\bar{r}\left(\gD_1\right)-\bar{r}\left(\gD_2\right)\right)\right]\geq\frac{1}{2}\textcolor{black}{\varsigma_P\left(\frac{V\left(\pi_1\right)-V\left(\pi_2\right)}{2}\right).}$$ where $\gD_1$ and $\gD_2$ are trajectory batches with size $D$ sampled under policies $\pi_1$ and $\pi_2$, and $\bar r$ returns to the averaged rewards over trajectories in one batch.
\end{definition}

This distinguishability characterizes the fundamental limit of using preferences over finite-size batches to recover the true preference between two policies because $\E\left[\varsigma_P(X)\right]\not=\varsigma_P(\E[X])$ in general. While the inequality always holds when $D\rightarrow \infty,$ it is possible that when $D$ is not large enough, the panel cannot reliably tell which policy is better from finite-size batches when the value functions of the two policies are too close, even with infinitely many panelists. $\epsilon^*_D$ defines this fundamental limit with a given $D.$

\subsection{Rate of Convergence}\label{rateofconvergence}

We present an upper bound on the convergence rate of \texttt{Par-\allowbreak S$^2$ZPO}.

\begin{theorem}\label{thm1}
    Assume time-homogeneous perturbation distance, i.e, $\mu_{t}=\mu$ for all $t$, and learning rates $\alpha_t=\Theta\left(\sqrt{{H}/{dt}}\right)$. After $T$ policy iterations under \texttt{Par-S$^2$ZPO}, if we sample policy $\vtheta_R$ from $\{\vtheta_t\}_{t=1}^T$ with probability $\sP\left(\vtheta_R=\vtheta_i\right)={\alpha_i}/{\sum_{t=1}^T\alpha_t}$, the following holds:$$\E\left[\left\|\nabla_\vtheta V\left(\pi_{\vtheta_R}\right)\right\|_{\gP_K}\right] = \Tilde{\gO}\left(\sqrt{\frac{Hd}{T}}+\mu d+\frac{\epsilon^*_DK}{\mu}+\frac{K}{\mu}\varsigma_P^{-1}\left(\sqrt{\frac{4}{N}}\right)\right).$$ 
\end{theorem}
Recall that $H$ is the planning horizon, $d$ is the parameter dimension, $T$ is the number of policy iterations, $\epsilon_D^*$ is the distinguishability, $K$ is the number of agents, $N$ is the number of batch pairs per policy iteration sampled by each agent, and $\varsigma_P\left(x\right)$ is the preference deviation function.  
Since $\nabla_\vtheta V\left(\pi_{\vtheta_t}\right)=0$ implies that the policy optimization reaches a stationary point, the inequality above provides an upper bound on the rate of convergence in terms of the number of policy iterations $T.$ The detailed proof of Theorem \ref{thm1} can be found in Section \ref{sectionproof}. The upper bound consists of three parts. 
\begin{itemize}[leftmargin=*]
    \item The first part $\Tilde{\gO}(\sqrt{{Hd}/{T}}+\mu d)$ is due to stochastic zeroth-order optimization, which is positively correlated to the perturbation distance $\mu.$ This is because by considering a perturbed policy that is $\mu$-distance away, the algorithm misses better policies that are closer than the perturbation distance. 

    \item The second part $\Tilde{\gO}({\epsilon^*_DK}/{\mu})$ is due to the distinguishability of human feedback defined in Definition \ref{HFlimitation} and the third part in the theorem $\Tilde{\gO}({K}\varsigma_P^{-1}(\sqrt{{4}/{N}})/{\mu})$ is a result of the majority vote error, both inverse proportional to the perturbation distance $\mu$. When the two policies are closer, their value difference becomes smaller. Since human panelists can only access a finite number of trajectory samples, the difference between the two values may become too small for the panelists to distinguish. 
\end{itemize}
From Theorem \ref{thm1},  by tuning the perturbation distance $\mu$, we can obtain the following upper bound. 

\begin{corollary}
Choosing the perturbation distance $\mu$ such that $\mu^2=\Theta({K}/{d}\max\{\varsigma_P^{-1}(\sqrt{{4}/{N}}),\epsilon^*_D\}),$ Theorem~\ref{thm1} becomes:
\begin{equation}
\E\left[\left\|\nabla_\vtheta V\left(\pi_{\vtheta_R}\right)\right\|_{\gP_K}\right] = \sqrt{d}\cdot\Tilde{\gO}\left(\sqrt{\frac{H}{T}}+\sqrt{\epsilon^*_DK}+\sqrt{K\varsigma_P^{-1}\left(\sqrt{\frac{4}{N}}\right)}\right).
\label{cor:ub}
\end{equation}
\end{corollary}

From this corollary, we can have some interesting observations.   Consider the case where $d$ is large and $K$ subsets are generated randomly and independently at each iteration.  Then, with a high probability, we have  
      $$\left\|\nabla_\vtheta V\left(\pi_{\vtheta_R}\right)\circ{\ve}_{\gI_k}\right\|_2\approx \left\|\nabla_\vtheta V\left(\pi_{\vtheta_R}\right)\circ{\ve}_{\gI_h}\right\|_2, \quad \forall k,h\in[K]$$ or  $$\left\|\nabla_\vtheta V\left(\pi_{\vtheta_R}\right)\right\|_{\gP_K}\approx \sqrt{K} \left\|\nabla_\vtheta V\left(\pi_{\vtheta_R}\right)\right\|_2.$$ Substituting this into Equation (\ref{cor:ub}), we have 
     \begin{equation}
     \E\left[\left\|\nabla_\vtheta V\left(\pi_{\vtheta_R}\right)\right\|_2\right]=\sqrt{d}\cdot\Tilde{\gO}\left(\sqrt{\frac{H}{TK}}+\sqrt{\epsilon^*_D}+\sqrt{\varsigma_P^{-1}\left(\sqrt{\frac{4}{N}}\right)}\right).\label{ub2}
\end{equation}
Recall that at each iteration, each agent samples $2ND$ trajectories for obtaining preferences. So over $T$ iterations, the algorithm samples $2NDKT$ trajectories in total. Let $M=2NDKT$ denote the total number of trajectories and let $M$ measure sample complexity. Then for given $M$ and $K,$ the algorithm can iterates $T=\frac{M}{2NDK}$ iterations. Substituting $M$ into the upper bound (\ref{ub2}) yields
\begin{align}
    \E\left[\left\|\nabla_\vtheta V\left(\pi_{\vtheta_R}\right)\right\|_2\right] = \sqrt{d}\cdot\Tilde{\gO}\left(\sqrt{\frac{2NDH}{M}}+\sqrt{\epsilon^*_D}+\sqrt{\varsigma_P^{-1}\left(\sqrt{\frac{4}{N}}\right)}\right).\label{ub3}
\end{align}
We can make two observations from this upper bound:
 \begin{itemize}[leftmargin=*]
     \item The upper bound is independent of $K.$ In other words, sample-wise, $K$-agent system is as efficient as a centralized, single-agent system ($K=1$), even though the single-agent system takes $K$ times more iterations under the same sample complexity. Our simulations will confirm this observation.  

     \item There is a tradeoff in choosing $D,$ the batch size, The first term $\sqrt{{2DH}/{M}}$ is positively correlated to $D$, whereas the second term $\sqrt{\epsilon_D^*}$ is inversely correlated to $D$ (with common human preference models, we have $\epsilon_D^*=\Tilde{O}(H/\sqrt{D})$~\cite{ZhaYin_25_Unknown}). A larger batch size $D$ will decrease the total training iterations while making each training iteration more accurate. There is also a tradeoff in choosing $N,$ similar to that of $D.$
 \end{itemize}

\subsection{Proof of Theorem \ref{thm1}}\label{sectionproof}

In this section, we prove Theorem~\ref {thm1} from a Lyapunov drift analysis framework. We start by analyzing the drift of the value function between successive policy iterations, i.e., $V\left(\pi_{\vtheta_{t+1}}\right) - V\left(\pi_{\vtheta_t}\right)$. 

Recall that by assumption, the value function $V\left(\pi_{\vtheta_t}\right)$ is $L$-smooth, so we can linearize the drift up to a second-order error, which is characterized by the following lemma:
\begin{lemma}\label{smoothapproximation}
    \textbf{(Lemma 7 in \cite{LiuKaiChe_18})} For any $L$-smooth function $f:\sR^d\rightarrow\sR$, and $\forall \vx,\vy\in\sR^d$, we have:
    $$
        |f\left(\vy\right)-f\left(\vx\right)-\left\langle\nabla f\left(\vx\right),\vy-\vx\right\rangle|\leq\frac{L}{2}\left\|\vy-\vx\right\|_2^2.
    $$
\end{lemma}
Substituting in the value function $V\left(\pi_{\vtheta_t}\right)$, we bound the drift as:
\begin{align*}
    &V\left(\pi_{\vtheta_{t+1}}\right)-V\left(\pi_{\vtheta_t}\right)\\ 
    \geq& \langle \nabla_\vtheta V\left(\pi_{\vtheta_t}\right), \vtheta_{t+1} - \vtheta_t \rangle - \frac{L}{2} \|\vtheta_{t+1} - \vtheta_t\|_2^2\\
    =& \alpha_t \langle \nabla_\vtheta V\left(\pi_{\vtheta_t}\right), \hat{{\vg}}_t \rangle - \frac{L\alpha_t^2}{2} \|\hat{{\vg}}_t\|_2^2 \\
    =& \alpha_t \sum_{k=1}^K\langle \nabla_\vtheta V\left(\pi_{\vtheta_t}\right), \hat{{O}}_{t,k}{\vv}_{t,k} \rangle - \frac{L\alpha_t^2}{2} \left\|\sum_{k=1}^K\hat{{O}}_{t,k}{\vv}_{t,k}\right\|_2^2,
\end{align*}
where the first equality uses the gradient ascent update and the second equality uses the definition of the gradient estimator $\hat{{\vg}}_t$ in Algorithm \ref{algorithm1}. Since the panelists' decision, i.e., $\hat{{O}}_{t,k}$, is related to the value difference between policies before and after perturbation, each inner product can be decomposed as follows:
\begin{align*}
    &\left\langle\nabla_\vtheta V\left(\pi_{\vtheta_t}\right),\hat{{O}}_{t,k}{\vv}_{t,k}\right\rangle\\
        =&\left\langle \nabla_\vtheta V\left(\pi_{\vtheta_t}\right), \operatorname{sign} \left[ V\left(\pi_{\vtheta_{t,k}'}\right) -V\left(\pi_{\vtheta_t}\right)\right] {\vv}_{t,k} \right\rangle\\
        & +\left\langle\nabla_\vtheta V\left(\pi_{\vtheta_t}\right), \left(\hat{{O}}_{t,k} - \operatorname{sign}\left[V \left(\pi_{\vtheta_{t,k}'} \right)-V\left(\pi_{\vtheta_t}\right)\right]\right){\vv}_{t,k}\right\rangle.
\end{align*}
The first term characterizes how the gradient direction of $\nabla_\vtheta V\left(\pi_{\vtheta_t}\right)$ aligns with the value difference sign between the perturbed and the original policy, and the second term quantifies the error of approximating the value difference sign with the majority vote from $N$ trajectories. The following two lemmas bound the conditional expectation of these two terms separately.

\begin{lemma}\label{1stbound}
Let $\gF_t$ be the filtration up to the $t$-th iteration, we have:
    \begin{align*}
        &\E\left[ \left. \left\langle\nabla_\vtheta V\left(\pi_{\vtheta_t}\right),\operatorname{sign}\left[ V\left(\pi_{\vtheta_{t,k}'}\right)-V\left(\pi_{\vtheta_t}\right) \right]{\vv}_{t,k}\right\rangle \right|\gF_t \right]\\
        \geq & \frac{1}{\sqrt{3}}\left\|\nabla_\vtheta V\left(\pi_{\vtheta_t}\right)\circ{\ve}_{\gI_k}\right\|_2-\mu_t L|\gI_k|.
    \end{align*}
\end{lemma}

\begin{lemma}\label{2ndbound}
The error of the majority vote is bounded as:
    \begin{align*}
        &\left| \E \left[ \left\langle \nabla_\vtheta V\left(\pi_{\vtheta_t}\right), \left( \hat{{O}}_{t,k} -\operatorname{sign} \left[ V\left( \pi_{\vtheta_{t,k}'} \right) -V\left(\pi_{\vtheta_t}\right) \right] \right) {\vv}_{t,k} \right \rangle| \gF_t \right] \right| \\& \leq\frac{2}{e^2}\left\|\nabla_\vtheta V\left(\pi_{\vtheta_t}\right)\circ{\ve}_{\gI_k}\right\|_2 + 2\left(\mu_t L|\gI_k|+\frac{\epsilon^*_D}{\mu_t}\right)+\frac{8}{\mu_t}\varsigma_P^{-1}\left(\sqrt{\frac{4}{N}}\right).
    \end{align*}
\end{lemma}
The proofs of Lemma~\ref{1stbound} and Lemma~\ref{2ndbound} are deferred to the appendix. Collecting these two bounds, we can lower bound the conditional expectation of the first term in the drift as follows:
\begin{align*}
    &\E\left[ \left. \sum_{k=1}^K \left\langle \nabla_\vtheta V\left(\pi_{\vtheta_t}\right), \hat{{O}}_{t,k}{\vv}_{t,k} \right\rangle \right| \gF_t \right] \\
    \geq& \left(\frac{1}{\sqrt{3}} - \frac{2}{e^2}\right) \sum_{k=1}^K\|\nabla_\vtheta V\left(\pi_{\vtheta_t}\right)\circ{\ve}_{\gI_k}\|_2\\
    &- \sum_{k=1}^K\left(3\mu L|\gI_k|+\frac{2\epsilon^*_D}{\mu}+\frac{8}{\mu}\varsigma_P^{-1}\left(\sqrt{\frac{4}{N}}\right)\right)\\
    \geq & \frac{1}{4} \|\nabla_\vtheta V\left(\pi_{\vtheta_t}\right)\|_{\gP_K} - 8\left(\mu Ld + \frac{\epsilon^*_D K}{\mu} + \frac{K}{\mu}\varsigma_P^{-1}\left(\sqrt{\frac{4}{N}}\right)\right),
\end{align*}
where the last step uses $\sum_{k=1}^K |\gI_k| = d$ since $\{\gI_k\}_{k=1}^K$ is a partition of coordinates. Next, we bound the second-order approximation error in the drift. Notice that for a different agent $k$, the perturbation vector ${\vv}_{t,k}$ is either $-1$ or $1$ on the coordinates in $\gI_k$ and $0$ at other entries. So for another agent $l$, the $\gI_l$ entries in vector ${\vv}_{t,k}$ should all be zero, which means the inner product of ${\vv}_{t,k}$ and ${\vv}_{t,l}$ should be $0$. Then, the second-order error is simply the following:
\begin{align*}
    \left\|\sum_{k=1}^K\hat{{O}}_{t,k}{\vv}_{t,k}\right\|_2^2 = \sum_{k=1}^K \hat{{O}}_{t,k}^2 \|{\vv}_{t,k}\|_2^2 = \sum_{k=1}^K |\gI_k| = d.
\end{align*}
Therefore, collecting the bounds of both terms in the drift analysis, we have:
\begin{align*}
    \E\left[V\left(\pi_{\vtheta_{t}}\right)-V\left(\pi_{\vtheta_{t+1}}\right)| \gF_t\right]
    \leq -\frac{\alpha_t}{4} \|\nabla_\vtheta V\left(\pi_{\vtheta_t}\right)\|_{\gP_K} + \frac{\alpha_t^2 L d}{2}\\
    + 8 \alpha_t \left(\mu Ld + \frac{\epsilon^*_D K}{\mu} + \frac{K}{\mu}\varsigma_P^{-1}\left(\sqrt{\frac{4}{N}}\right)\right).
\end{align*}
Now, we take the expectation over $\gF_t$ and then a telescoping sum over $t=1$ to $T$, we have:
\begin{align*}
    \E\left[V\left(\pi_{\vtheta_{1}}\right) - V\left(\pi_{\vtheta_{T+1}}\right)\right] 
    \leq - \sum_{t=1}^{T}\alpha_t \frac{\E\left[ \|\nabla_\vtheta V\left(\pi_{\vtheta_t}\right)\|_{\gP_K} \right]}{4} \\
    + \sum_{t=1}^{T} 8\alpha_t \left(\mu Ld + \frac{\epsilon^*_D K}{\mu} + \frac{K}{\mu}\varsigma_P^{-1}\left(\sqrt{\frac{4}{N}}\right)\right) + \sum_{t=1}^T \alpha_t^2 \frac{Ld}{2}.
\end{align*}
Then, we can bound the gradient norm in Theorem~\ref{thm1} as follows:
\begin{align*}
    \E\left[\|\nabla_\vtheta V\left(\pi_{\vtheta_R}\right)\|_{\gP_K}\right]
    =& \frac{\sum_{t=1}^T \alpha_t \E\left[ \|\nabla_\vtheta V\left(\pi_{\vtheta_t}\right)\|_{\gP_K} \right]}{\sum_{t=1}^T \alpha_t}\\
    \leq & \frac{4 \E\left[V\left(\pi_{\vtheta_{1}}\right) - V\left(\pi_{\vtheta_{T+1}}\right)\right]}{\sum_{t=1}^T \alpha_t} + \frac{2 Ld \sum_{t=1}^T \alpha_t^2}{\sum_{t=1}^T \alpha_t}\\
    &+32\left(\mu Ld + \frac{\epsilon^*_D K}{\mu} + \frac{K}{\mu}\varsigma_P^{-1}\left(\sqrt{\frac{4}{N}}\right)\right).
\end{align*}
By choosing $\alpha_t=\Theta\left(\sqrt{{H}/{dt}}\right)$, then $\sum_{t=1}^T\alpha_t=\Theta\left(\sqrt{{HT}/{d}}\right)$ and $\sum_{t=1}^T\alpha_t^2=\Theta\left({H\log T}/{d}\right)$. Note $V\left(\pi_{\vtheta_{1}}\right) - V\left(\pi_{\vtheta_{T+1}}\right) \leq H$, we obtain
\begin{align*}
    \E\left[\|\nabla_\vtheta V\left(\pi_{\vtheta_R}\right)\|_{\gP_K}\right] = \Tilde{\gO}\left(\sqrt{\frac{Hd}{T}}+\mu d+\frac{\epsilon^*_DK}{\mu}+\frac{K}{\mu}\varsigma_P^{-1}\left(\sqrt{\frac{4}{N}}\right)\right),
\end{align*}
where $\Tilde{\gO}\left(\cdot\right)$ suppresses logarithmic dependency. This proves our main theorem.

\section{Experiments}

This section presents the performance of \texttt{Par-S$^2$ZPO} in four Mujoco environments~\cite{gymnasium}: Half Cheetah, Hopper, Swimmer, and Walker2D. In Mujoco environments, RL policies take the robots' positions and postures (observations) as input, and output the forces and torques to control the robots to accomplish the tasks. The performances are measured by the accumulated rewards defined by the environments. 

All policies are parameterized by neural networks (actor networks) with two hidden layers of size 64, while input and output dimensions depend on the environment. Each actor is first pretrained using PPO~\cite{SchWolDha_17}, providing a stable and stochastic initialization for further zeroth-order fine-tuning. The preference model of each panelist is a linear link function
\begin{equation*}
\sigma(x)=\max\{0,\min\{ax+0.5,1\}\}, \quad a=0.01.
\end{equation*} Each agent is associated with a panel of $P=100$ panelists. For each query, the panel samples one pair of trajectory batches ($N=1$), collects individual preferences, and returns the majority vote. The underlying rewards and link function are unknown to the agents. Each experiment is initialized with learning rate $\alpha=0.01$ and perturbation magnitude $d=0.05$. At each policy iteration, after receiving feedback from all agents, a server-side panel (identical to agent panels) evaluates whether the updated policy would be preferred over the current one. The update is accepted only if preferred; otherwise, it is rejected. If three consecutive rejections occur, both $\alpha$ and $d$ are halved. If the updated policy is accepted, Adam~\cite{adam} with gradient clipping was then used. 
When evaluating the performance of a policy at each iteration, we use $20$ new episodes.

Our experiments focus on the following aspects: (1) the impact of batch size $D,$ (2) the impact of the number of agents $K,$ (3) binary versus Gaussian perturbation,  and (4) performance comparison with the FedAvg-type of federated learning framework. 

\subsubsection*{Batch size $D$ selection}

The upper bound (\ref{ub3}) reveals a tradeoff in batch size $D$. The different choices of $D$'s can be explained by the distinguishability $\epsilon_D^*$, which is a decreasing function of $D$. Under a fixed sample complexity, when $\epsilon_D^*$ is already relatively small, continuing to increase $D$ can result in a decrease in the total training iteration, which is wasteful and impacts the training performance.

\begin{figure}[h]
    \centering
    \begin{subfigure}[b]{0.48\textwidth}
        \centering
        \includegraphics[width=0.48\textwidth]{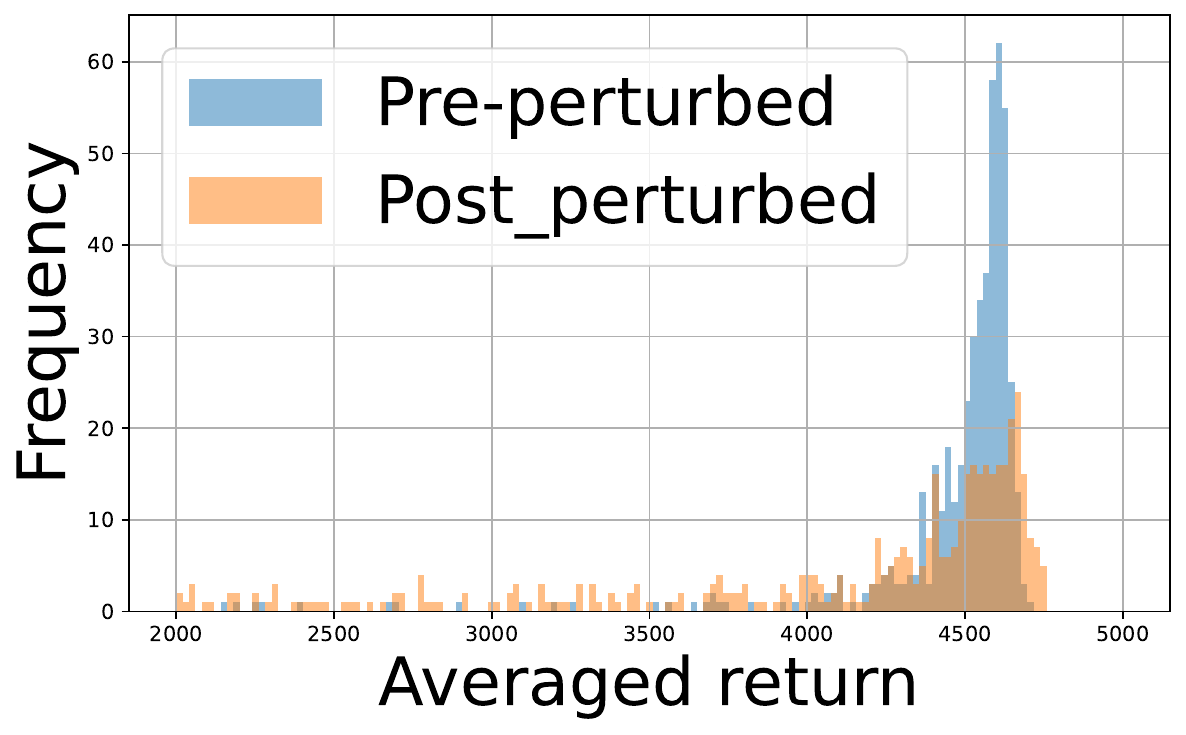}
        \includegraphics[width=0.48\textwidth]{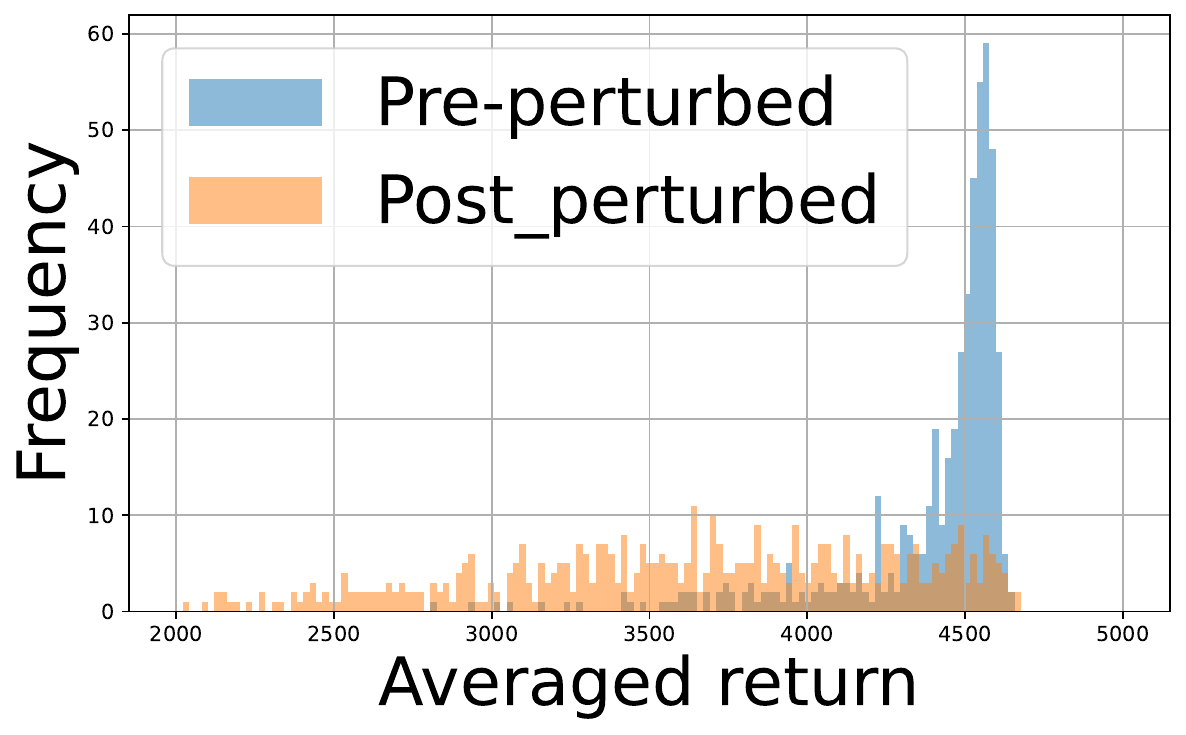}
        \caption{Half Cheetah}
    \end{subfigure}
    \begin{subfigure}[b]{0.48\textwidth}
        \centering
        \includegraphics[width=0.48\textwidth]{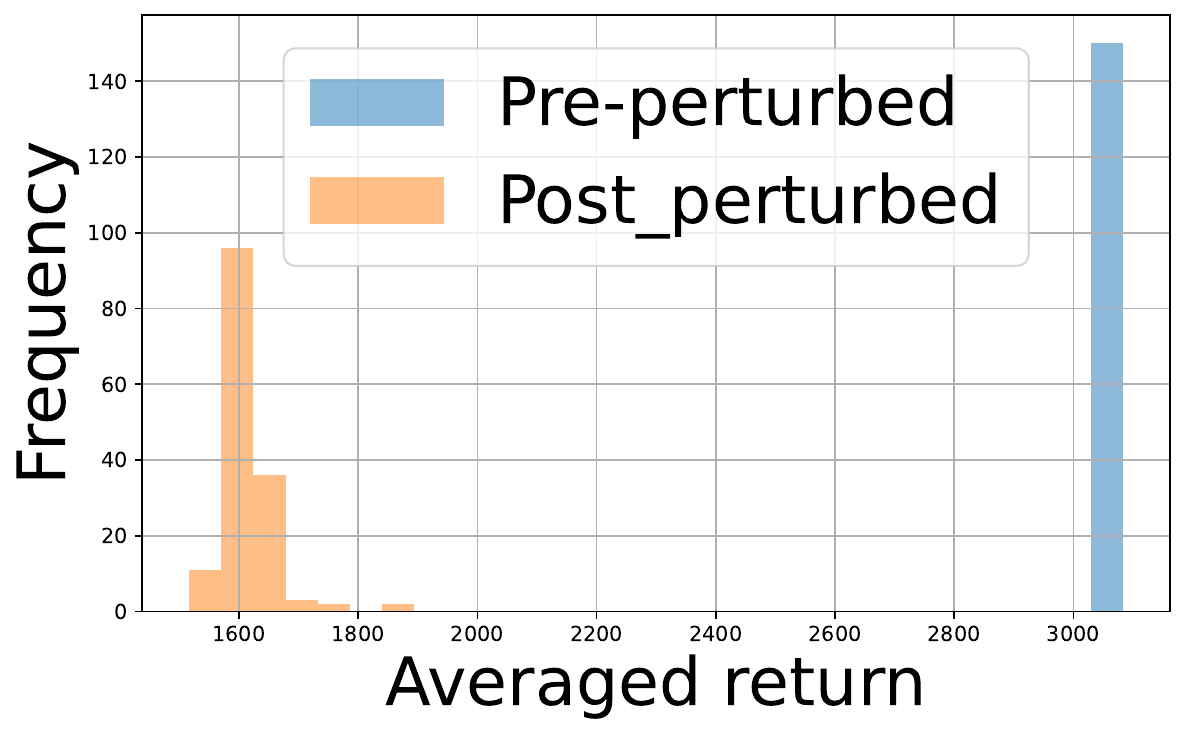}
        \includegraphics[width=0.48\textwidth]{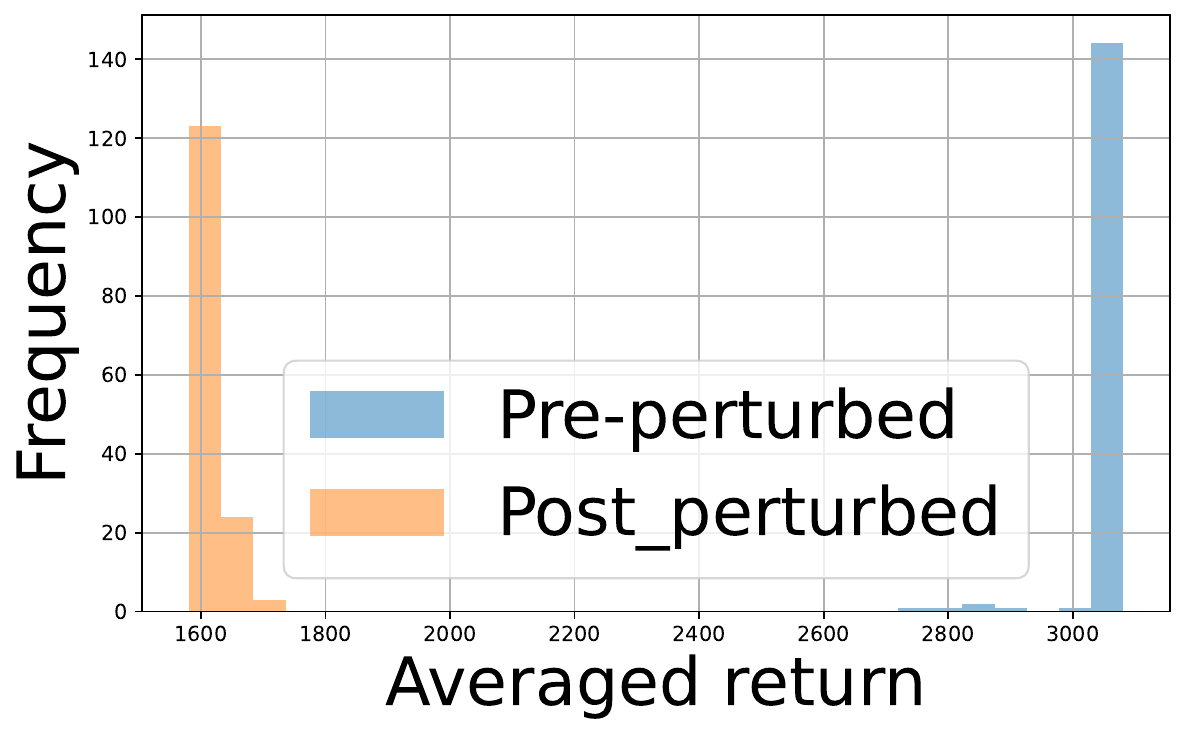}
        \caption{Hopper}
    \end{subfigure}
    \begin{subfigure}[b]{0.48\textwidth}
        \centering
        \includegraphics[width=0.48\textwidth]{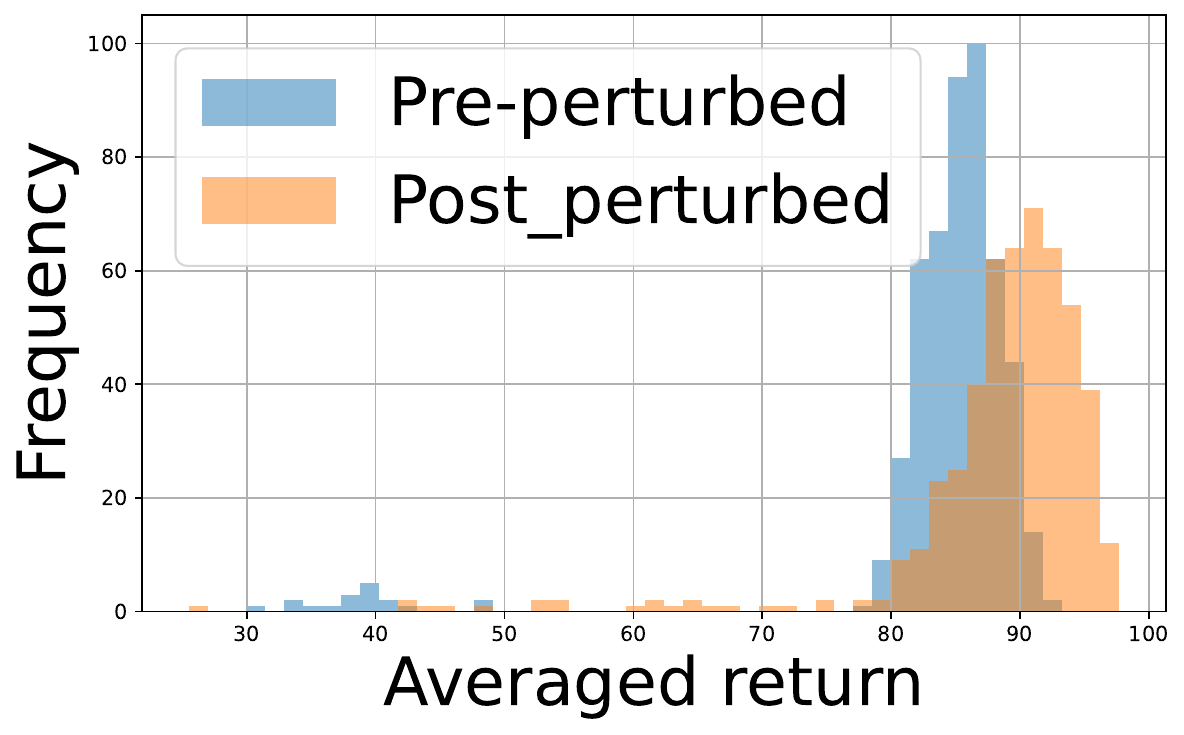}
        \includegraphics[width=0.48\textwidth]{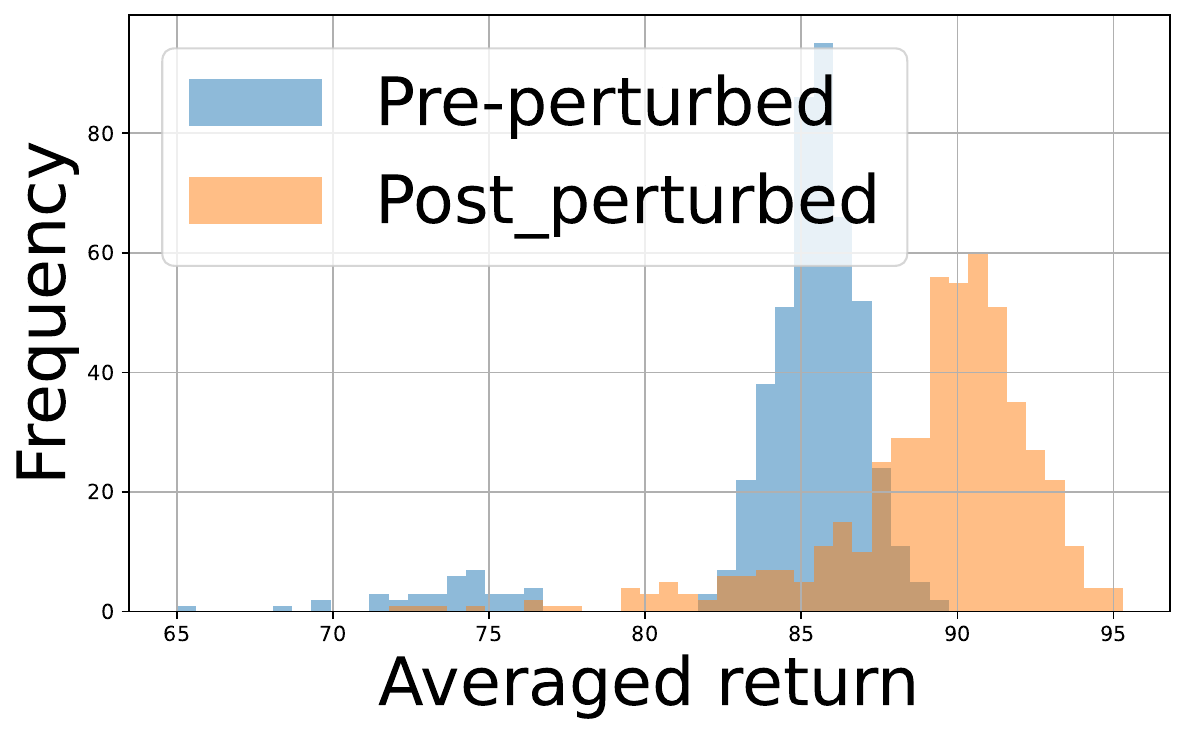}
        \caption{Swimmer}
    \end{subfigure}
    \begin{subfigure}[b]{0.48\textwidth}
        \centering
        \includegraphics[width=0.48\textwidth]{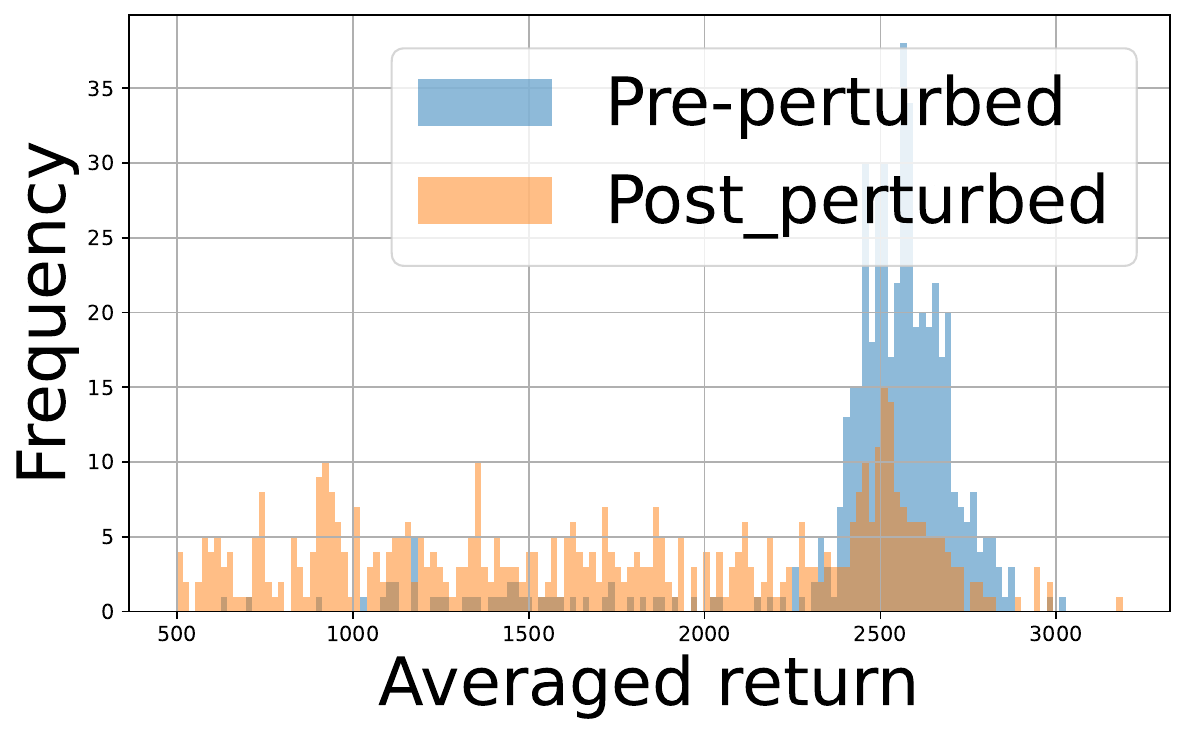}
        \includegraphics[width=0.48\textwidth]{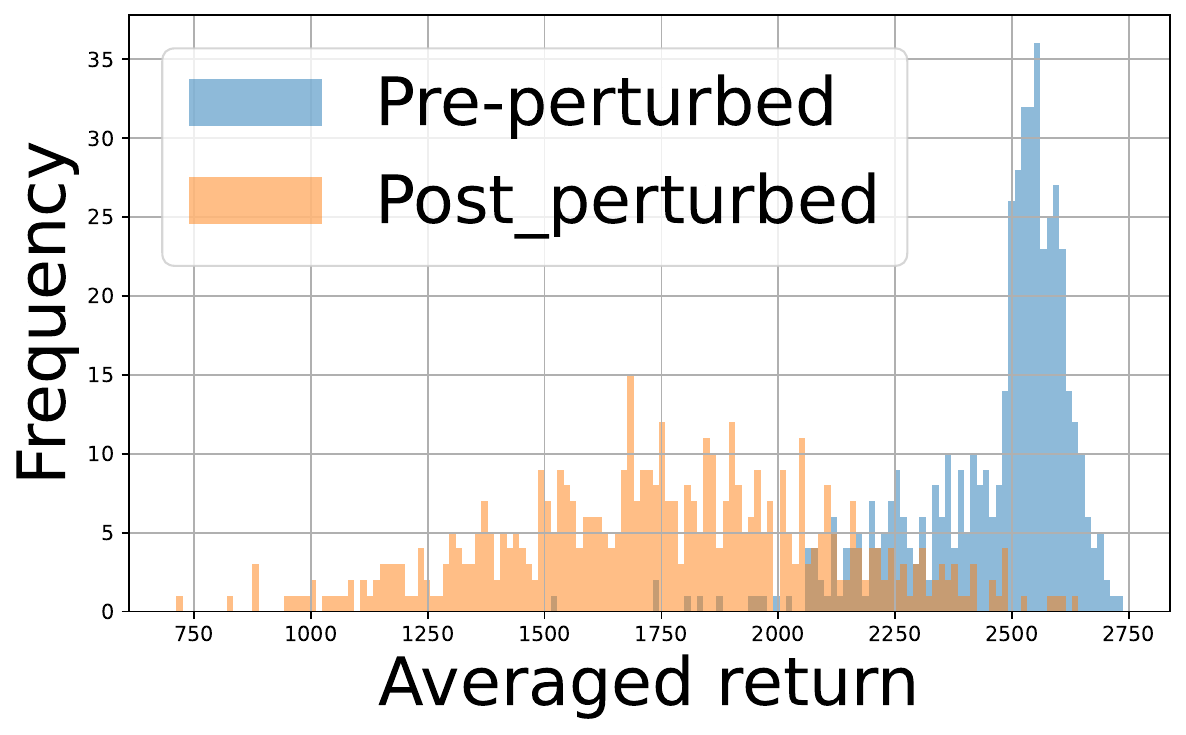}
        \caption{Walker 2D}
    \end{subfigure}
    \caption{Histograms of sample batches' average returns from pre-perturbed and post-perturbed policies under different batch size $D$. For each environment, the first histogram shows the case when $D=1$, the second shows the case when $D=4$.}
    \label{fig:d_com}
\end{figure}

In Figure~\ref{fig:d_com}, we plotted the empirical distributions of the average returns of $D$ trajectories of the policies before and after perturbation. 
For Half Cheetah, Swimmer and Walker 2D, we sampled $500$ batches per policy, and for Hopper, we sampled $150$ batches per policy.

We can observe that for Hopper, the two policies are already well separated at $D=1$, so $\epsilon_D^*$ is sufficiently small even with $D=1$. For Swimmer, increasing $D$ reduces the overlap between the two distributions, but the separation is clear even when $D=1$. Therefore, increasing $D$ provides no additional benefit and only reduces the number of policy iterations. For Half Cheetah and Walker 2D, the empirical distributions of the policies before and after perturbation have significant overlap, which decreases significantly when $D$ increases from 1 to 4. Therefore, increasing $D$ improves distinguishability significantly, which leads to the tradeoff in the upper bound (\ref{ub3}). In the following experiments, for Hopper and Swimmer, we select $D=1$, and for Half Cheetah and Walker 2D, we select $D=4$.

\subsubsection*{Agent size $K$ study}

In this experiment, we studied the performance of \texttt{Par-S$^2$ZPO} with different values of $K=1,\ 5,\ 10,\ 15$ under the sample complexity.  
i.e., the training iteration decreases proportionally with the increase of $K$. 
We repeated the training $20$ times for each $K$. The training performances with standard deviations are shown in Table \ref{tab:k_com}. 
From the table, we can observe that after considering the standard deviations, the performance remains the same across different $K$'s, which is consistent with Section \ref{rateofconvergence}.

\subsubsection*{Binary versus Gaussian perturbations}
Our binary perturbation leads to low communication, computation, and memory complexity. In this experiment, we compared the performances of the binary and Gaussian perturbations. From each environment, we choose the agent size $K=1$ to compare the two perturbations. We set the training iteration $T=150$ for Half Cheetah, Swimmer, and Walker 2D, and $T=30$ for Hopper since the training in Hopper converges faster.  The comparison is shown in the first two columns of Table~\ref{tab:k_com}, which indicates that the two perturbation methods have similar performance. This justifies our choice of binary perturbation. 

\begin{table}[h]
\centering
\small
\begin{tabular}{cccccc}
\toprule
Environments & $K=1$ (G) & $K=1$ (B) & $K=5$ (B) & $K=10$ (B) & $K=15$ (B) \\
\midrule
Half Cheetah 
& $4375.40\pm30.90$ 
& $4394.77\pm27.97$ 
& $4398.68\pm31.66$ 
& $4422.25\pm24.62$ 
& $4424.69\pm29.49$ \\

Hopper 
& $2557.88\pm119.43$ 
& $2624.65\pm78.81$ 
& $2658.00\pm76.46$ 
& $2594.85\pm102.66$ 
& $2583.12\pm84.68$ \\

Swimmer 
& $94.85\pm4.53$ 
& $91.37\pm1.74$ 
& $94.57\pm1.96$ 
& $92.59\pm1.32$ 
& $91.80\pm1.37$ \\

Walker2D 
& $2376.34\pm58.49$ 
& $2433.64\pm54.93$ 
& $2649.10\pm42.00$ 
& $2552.52\pm60.18$ 
& $2651.27\pm50.74$ \\

\bottomrule
\end{tabular}
\caption{Performance under different $K$ and perturbation types (\textit{G}aussian/\textit{B}inary) with the same sample complexity.}
\label{tab:k_com}
\end{table}

\subsubsection*{Performance comparison with FedAvg}

\begin{figure}[h]
    \centering
    \begin{subfigure}[b]{0.24\textwidth}
        \centering
        \includegraphics[width=\textwidth]{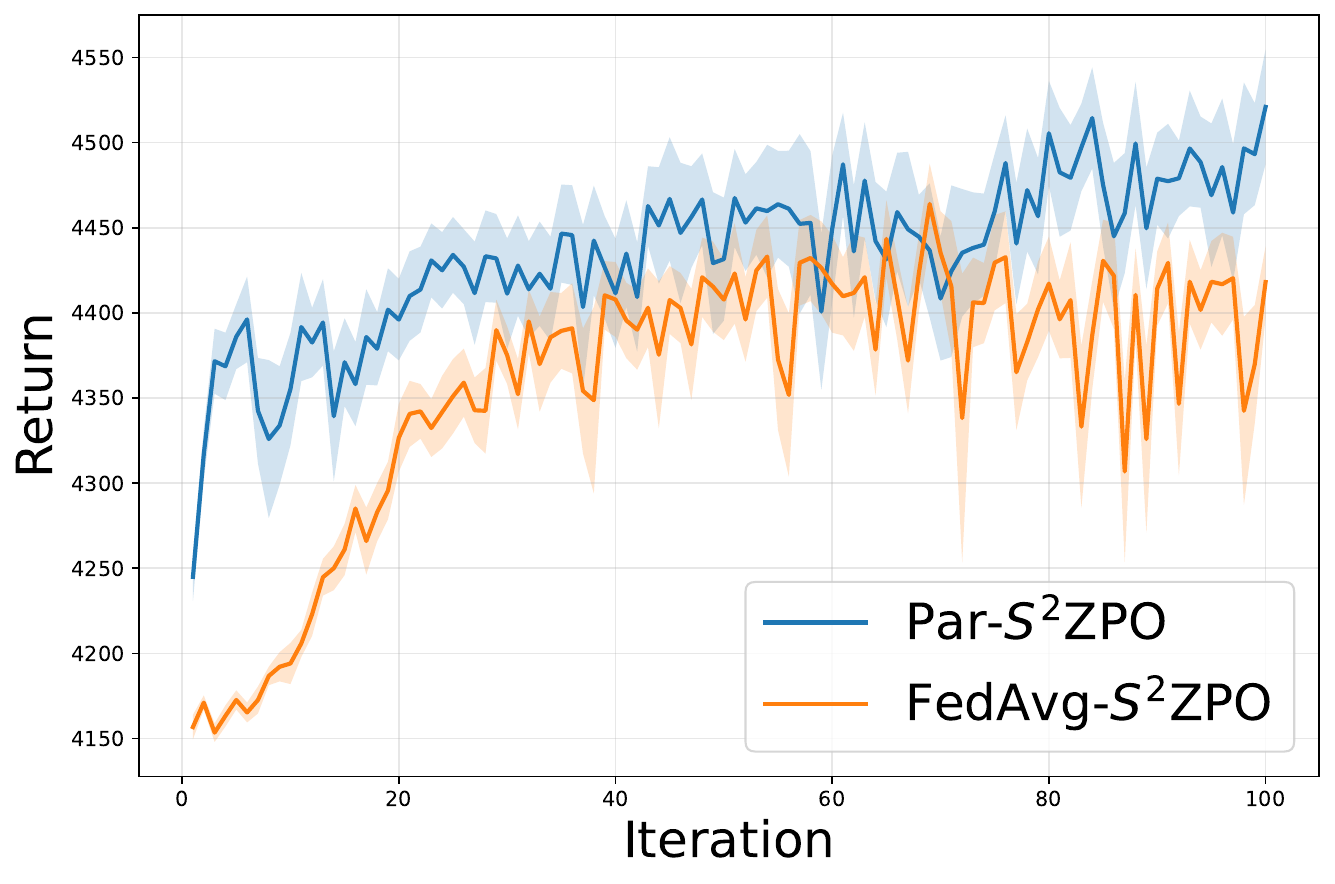}
        \caption{Half Cheetah}
    \end{subfigure}
    \begin{subfigure}[b]{0.24\textwidth}
        \centering
        \includegraphics[width=\textwidth]{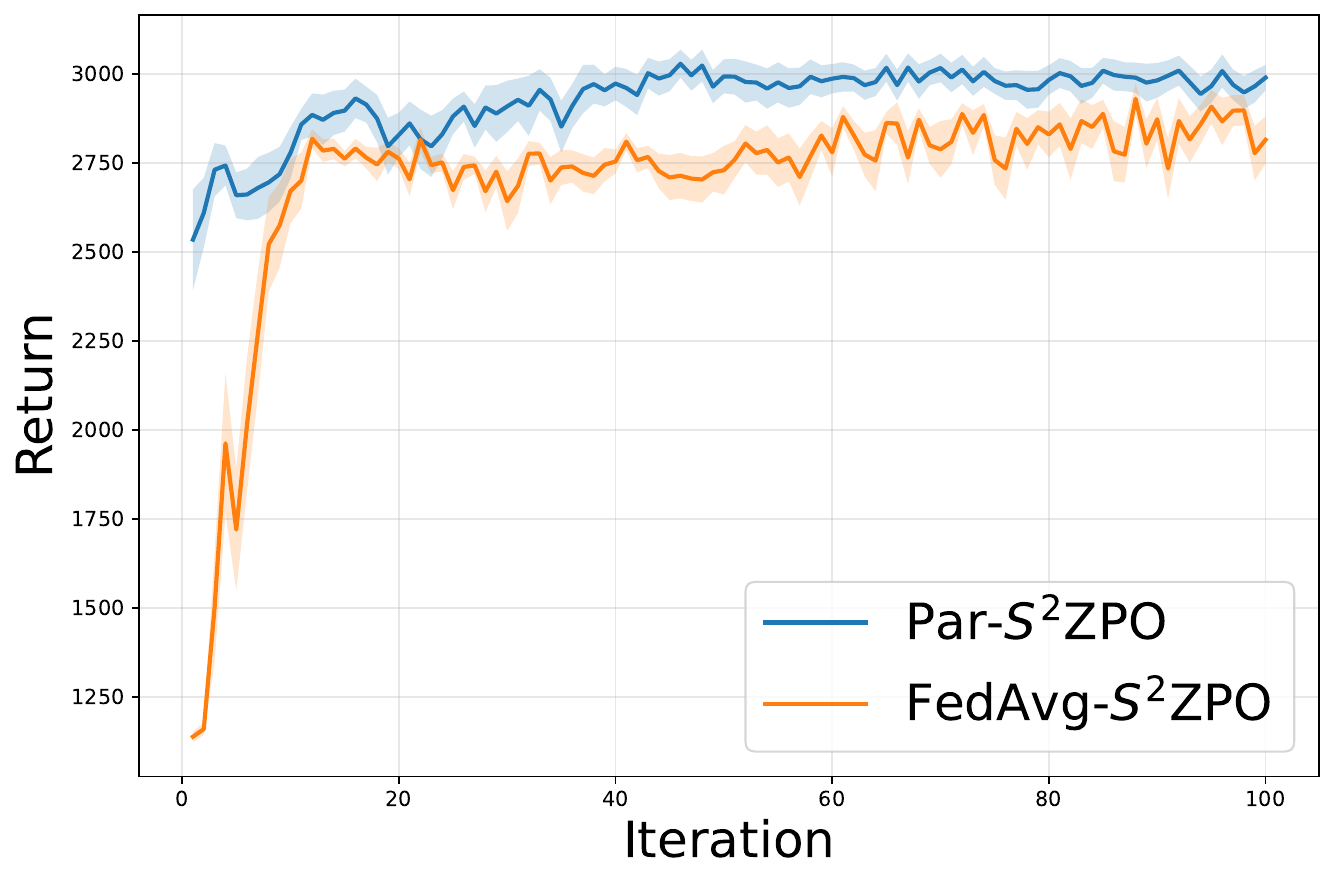}
        \caption{Hopper}
    \end{subfigure}
    \begin{subfigure}[b]{0.24\textwidth}
        \centering
        \includegraphics[width=\textwidth]{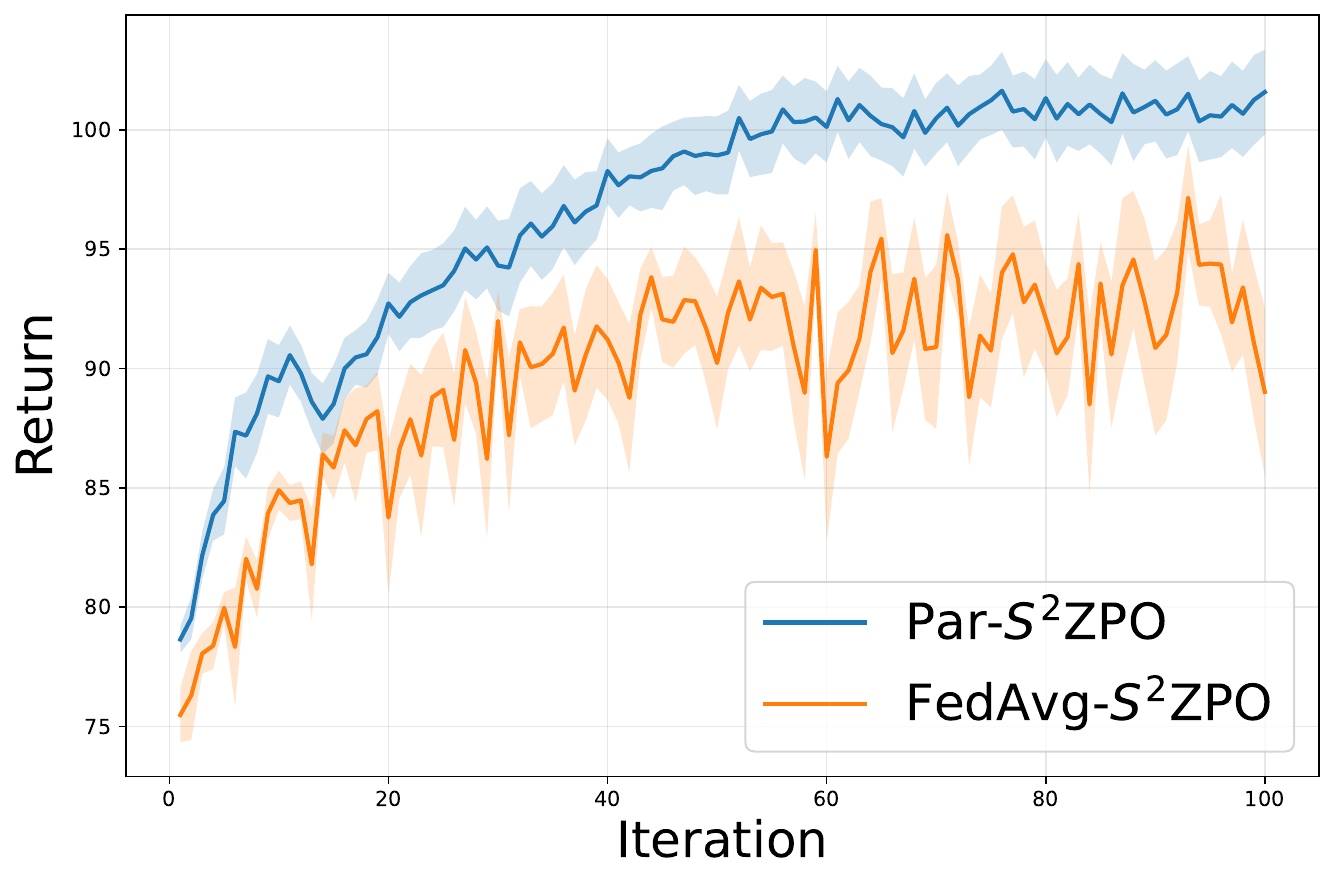}
        \caption{Swimmer}
    \end{subfigure}
    \begin{subfigure}[b]{0.24\textwidth}
        \centering
        \includegraphics[width=\textwidth]{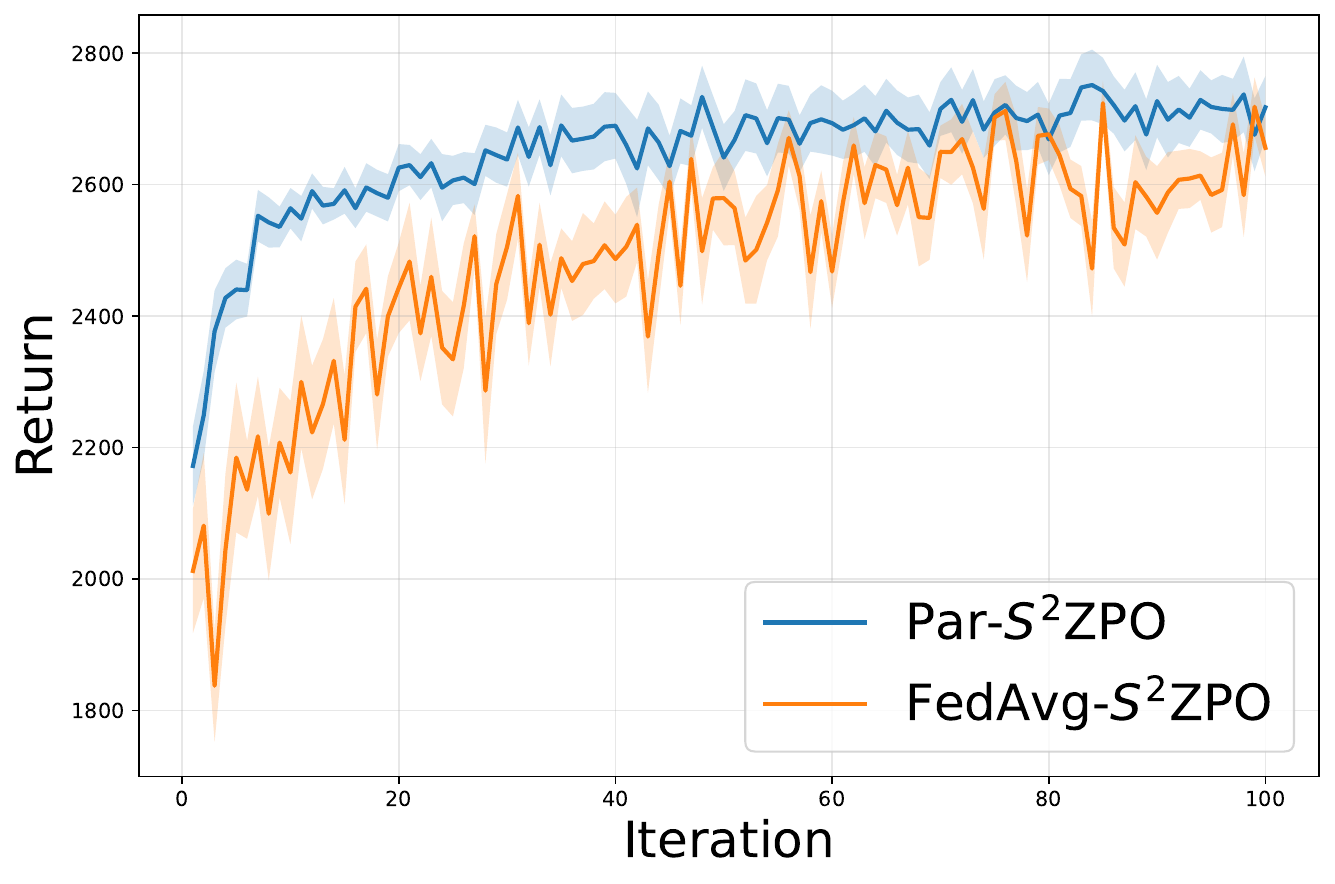}
        \caption{Walker 2D}
    \end{subfigure}
    \caption{\texttt{Par-S$^2$ZPO} versus \texttt{FedAvg-S$^2$ZPO}}
    \label{comparison}
\end{figure}

In this experiment, we compared \texttt{Par-S$^2$ZPO} with \texttt{FedAvg-S$^2$ZPO}. We used $K=5$ distributed agents. Each method is run for 20 independent trials, and results with standard errors are reported in Figure~\ref{comparison}. As shown in Figure~\ref{comparison}, \texttt{Par-S$^2$ZPO} consistently outperforms \texttt{FedAvg\allowbreak-S$^2$ZPO} across all environments. \texttt{Par-S$^2$ZPO} partitions the model into $K=5$ components and independently selects perturbation directions, effectively choosing from $2^K=32$ potential updates. This enables finer and more efficient policy improvement. For \texttt{FedAvg-S$^2$ZPO}, each agent selects its locally preferred update, and the server averages these updates. This aggregation leads to a coarser update compared to the combinatorial selection in \texttt{Par-S$^2$ZPO}.

Besides a better performance, \texttt{Par-S$^2$ZPO} is more communication- and memory-efficient. In each iteration, \texttt{FedAvg-S$^2$ZPO} requires transmitting $K$ full-network perturbations, whereas \texttt{Par-S$^2$ZPO} transmits only one perturbation split across $K$ partitions, reducing communication cost by a factor of $K$. Under \texttt{FedAvg-S$^2$ZPO}, each agent must store two full actor networks (original and perturbed) for inference, while \texttt{Par-S$^2$ZPO} requires only one full network and the partial network corresponding to its perturbation partition. For $K=5$, this reduces storage by approximately $40\%$.

\section{Conclusion}

In this paper, we proposed \texttt{Par-$S^2$ZPO} for Federated RLHF. The algorithm is communication-, computation-, and memory-efficient. The upper bound on the rate of convergence sheds light on the algorithm's efficiency.  Our empirical results confirmed the observations from our theoretical analysis. 

\bibliographystyle{abbrv}
\bibliography{inlab-refs.bib}

\appendix
\section*{Appendix}

\section{Proof of Lemma \ref{1stbound}}\label{proof_1stbound}
When the sampled perturbation vector $\vv_{t,k}$ aligns well with the positive/negative gradient direction on partition $\gI_k$, i.e., when $|\langle\nabla_\vtheta V\left(\pi_{\vtheta_t}\right),\vv_{t,k}\rangle|$ is large, the value difference $V(\pi_{\vtheta_{t,k} + \mu \vv_{t,k}}) - V(\pi_{\vtheta_t})$ on this direction will be larger than the first-order approximation error, which implies the sign of the policy value difference is the same as the sign of $\langle\nabla_\vtheta V\left(\pi_{\vtheta_t}\right),\vv_{t,k}\rangle$. Therefore, define $\gE_{t,k}$ to be the event where $\vv_{t,k}$ does not align with the gradient well, i.e.,
\begin{align*}
    \gE_{t,k}:=\{{\vv}_{t,k}:\left|\left\langle\nabla_\vtheta V\left(\pi_{\vtheta_t}\right),{\vv}_{t,k}\right\rangle\right| \leq \frac{L\mu_{t}}{2}\left\|{\vv}_{t,k}\right\|_2^2\}.
\end{align*}
Then, on its complement $\gE_{t,k}^{\mathsf{c}}$, the sign of the first-order approximation should be the same as the value function difference. Indeed, due to the smoothness of $V(\cdot)$ in Lemma~\ref{smoothapproximation}, when $\langle \nabla_\vtheta V(\pi_{\vtheta_t}), \mu_{t}{\vv}_{t,k} \rangle>0$, we have on $\gE_{t,k}^{\mathsf{c}}$ that:
\begin{align}
    V(\pi_{\vtheta_{t,k} + \mu \vv_{t,k}})-V(\pi_{\vtheta_t})
    \geq & \left\langle \nabla_\vtheta V(\pi_{\vtheta_t}), \mu_{t}{\vv}_{t,k} \right\rangle \nonumber
    - \frac{L}{2}\left\|\mu_{t}{\vv}_{t,k}\right\|_2^2\\
    \geq & \frac{L\mu_{t}^2}{2}\left\|{\vv}_{t,k}\right\|_2^2 - \frac{L}{2}\left\|\mu_{t}{\vv}_{t,k}\right\|_2^2 = 0. \nonumber
\end{align}
Thus, $\sign[V(\pi_{\vtheta_{t,k} + \mu \vv_{t,k}})-V(\pi_{\vtheta_t})] = \sign[\langle \nabla_\vtheta V(\pi_{\vtheta_t}), \mu_{t}{\vv}_{t,k} \rangle] = 1$. The same argument applies to where $\langle \nabla_\vtheta V(\pi_{\vtheta_t}), \mu_{t}{\vv}_{t,k} \rangle<0$. As a consequence, we have on $\gE_{t,k}^{\mathsf{c}}$ that:
\begin{align}\label{smooth2}
    &\left\langle\nabla_\vtheta V\left(\pi_{\vtheta_t}\right),\sign\left[V\left(\pi_{\vtheta_{t,k}'}\right)-V\left(\pi_{\vtheta_t}\right)\right]{\vv}_{t,k}\right\rangle \nonumber\\
    =& \sign[\langle \nabla_\vtheta V(\pi_{\vtheta_t}), \mu_{t}{\vv}_{t,k} \rangle]\left\langle \nabla_\vtheta V\left(\pi_{\vtheta_t}\right),{\vv}_{t,k}\right\rangle \nonumber\\
    = & \left|\left\langle\nabla_\vtheta V\left(\pi_{\vtheta_t}\right),{\vv}_{t,k}\right\rangle\right|.
\end{align}
Therefore, we analyze its conditional expectation as follows:
\begin{align*}
    &\E\left[\left. \left\langle \nabla_\vtheta V\left(\pi_{\vtheta_t}\right), \sign\left[ V\left(\pi_{\vtheta_{t,k}'}\right)-V\left(\pi_{\vtheta_t}\right) \right]{\vv}_{t,k} \right\rangle \right| \gF_t\right]\\
    =& \E_{{\vv}_{t,k}}\left[ \left\langle \nabla_\vtheta V\left(\pi_{\vtheta_t}\right), \sign\left[ V\left(\pi_{\vtheta_{t,k}'}\right) -V\left(\pi_{\vtheta_t}\right) \right] {\vv}_{t,k}  \right\rangle \mathbbm{1}_{\gE_{t,k}}\right]\\
    & + \E_{{\vv}_{t,k}}\left[ \left\langle \nabla_\vtheta V\left(\pi_{\vtheta_t}\right), \sign\left[ V\left(\pi_{\vtheta_{t,k}'}\right) -V\left(\pi_{\vtheta_t}\right) \right] {\vv}_{t,k}  \right\rangle \mathbbm{1}_{\gE_{t,k}^{\mathsf{c}}} \right]\\
    =& \E_{{\vv}_{t,k}}\left[ \left|\left\langle\nabla_\vtheta V\left(\pi_{\vtheta_t}\right),{\vv}_{t,k}\right\rangle\right| \right]\\ 
    & - \E_{{\vv}_{t,k}}\left[ \left\langle \nabla_\vtheta V\left(\pi_{\vtheta_t}\right), \sign[\langle \nabla_\vtheta V(\pi_{\vtheta_t}), \mu_{t}{\vv}_{t,k} \rangle] {\vv}_{t,k}  \right\rangle \mathbbm{1}_{\gE_{t,k}} \right] \\
    &+ \E_{{\vv}_{t,k}}\left[ \left\langle \nabla_\vtheta V\left(\pi_{\vtheta_t}\right), \sign\left[ V\left(\pi_{\vtheta_{t,k}'}\right) -V\left(\pi_{\vtheta_t}\right) \right] {\vv}_{t,k}  \right\rangle \mathbbm{1}_{\gE_{t,k}} \right]\\
    \geq & \E_{{\vv}_{t,k}}\left[ \left|\left\langle\nabla_\vtheta V\left(\pi_{\vtheta_t}\right),{\vv}_{t,k}\right\rangle\right| \right] - 2\E_{{\vv}_{t,k}}\left[ \left|\left\langle \nabla_\vtheta V\left(\pi_{\vtheta_t}\right), {\vv}_{t,k} \right\rangle\right| \mathbbm{1}_{\gE_{t,k}} \right]\\
    \geq & \E_{{\vv}_{t,k}}\left[ \left|\left\langle\nabla_\vtheta V\left(\pi_{\vtheta_t}\right),{\vv}_{t,k}\right\rangle\right|\right]-\mu_{t}L\E_{{\vv}_{t,k}}\left[\left\|{\vv}_{t,k}\right\|_2^2\right],
\end{align*}
where the second inequality uses equation~\ref{smooth2} on $\gE_{t,k}^{\mathsf{c}}$, and the last inequality uses the definition of $\gE_{t,k}$. Since each nonzero entry of ${\vv}_{t,k}$ is uniformly sampled from $\{-1,+1\}$, we have $\E[\|{\vv}_{t,k}\|_2^2]=|\gI_k|$. Thus, we obtain:
\begin{align}\label{1stbound1}
    &\E\left[\left. \left\langle \nabla_\vtheta V\left(\pi_{\vtheta_t}\right), \sign \left[ V\left(\pi_{\vtheta_{t,k}'}\right)-V\left(\pi_{\vtheta_t}\right) \right]{\vv}_{t,k} \right\rangle \right| \gF_t\right] \nonumber \\
    \geq & \E_{{\vv}_{t,k}}\left[ \left|\left\langle\nabla_\vtheta V\left(\pi_{\vtheta_t}\right),{\vv}_{t,k}\right\rangle\right|\right] - \mu_{t}L |\gI_k|.
\end{align}
The first term above is the absolute value of the sum of a sequence of Rademacher random variables, which can be characterized with the following lemma:
\begin{lemma}\label{Khintchine}
    Let $v_1,v_2,\dots,v_n$ be $n$ independent Rademacher random variables uniformly sampled from $\{-1,+1\}$, then for any sequence $a_1,a_2,\dots,a_n$, the following holds, 
    $$
        \frac{1}{\sqrt{3}} \left( \sum_{i=1}^n a_i^2 \right)^{\frac{1}{2}} \leq \E \left[ \left| \sum_{i=1}^n v_ia_i \right| \right] \leq \left( \sum_{i=1}^n a_i^2 \right)^{\frac{1}{2}}.
    $$
\end{lemma}
According to Lemma \ref{Khintchine}, let $a_i = \partial V(\pi_\vtheta)/ \partial \theta_i$ if $i\in \gI_k$ and $0$ otherwise, and $v_i$ be the $i$-th entry of the perturbation vector $v_{t,k}$, we have:
\begin{align*}
    \E_{{\vv}_{t,k}}\left[ \left|\left\langle \nabla_\vtheta V\left(\pi_{\vtheta_t}\right),{\vv}_{t,k}\right\rangle\right|\right] \geq \frac{1}{\sqrt{3}} \|\nabla_\vtheta V\left(\pi_{\vtheta_t}\right)\circ {\ve}_{\gI_k}\|_2.
\end{align*}
Substituting this lower bound back into equation~\ref{1stbound1}, we complete the proof of Lemma \ref{1stbound}.

\subsection{Proof of Lemma~\ref{Khintchine}}
Define a random variable $X = \sum_{i=1}^n v_i a_i$. For the upper bound, we have by Cauchy-Schwartz inequality:
\begin{align*}
    \E[|X|] & \leq \left( \E[X^2] \right)^{\frac{1}{2}}
    =\left(  \sum_{i=1}^n a_i^2 \E\left[v_i^2\right]  + \sum_{i\neq j} a_i a_j \E\left[v_i\right]  \E\left[v_j \right] \right)^{\frac{1}{2}}\\
    &=\left( \sum_{i=1}^n a_i^2 \right)^{\frac{1}{2}},
\end{align*} 
where the first equality uses independence among $v_1$ to $v_n$, and the second equality uses $E[v_i] = 0$ and $E[v_i^2]=1$ for any $i$. To derive the lower bound, according to H\"older's inequality, we have, 
$$
    \E[X^2]=\E\left[ |X|^{\frac{2}{3}} (X^4)^\frac{1}{3} \right] \leq \E[|X|]^\frac{2}{3} \E\left[ X^4 \right]^\frac{1}{3}.
$$ 
Then, we can lower bound $E[|X|]$ by its higher moments: 
$$
    \E[|X|]\geq\frac{\left(\E\left[ X^2 \right]\right)^\frac{3}{2}}{\left(\E\left[ X^4 \right]\right)^\frac{1}{2}} = \frac{\left( \sum_{i=1}^n a_i^2 \right)^{\frac{3}{2}}}{\E\left[ X^4 \right]^\frac{1}{3}},
$$
For the denominator, we first have \begin{align*}
    \E [X^4] &=  \sum_{i=1}^n a_i^4 \E\left[v_i^4\right] + 6\sum_{i<j}a_i^2 a_j^2 \E\left[v_i^2\right] \E\left[ v_j^2 \right]\\
    &= \sum_{i=1}^n a_i^4 + 6\sum_{i<j}a_i^2a_j^2,
\end{align*}
where in the first inequality we used the independence and $\E[v_i]=0$ again, so the odd moments disappear. To match the numerator:
$$
    \E [X^4] \leq 3\left( \sum_{i=1}^na_i^4 + 2\sum_{i<j} a_i^2a_j^2 \right) \leq 3\left( \sum_{i=1}^n a_i^2 \right)^{2}.
$$

Thus, we have, $$\E[|X|]\geq\frac{\left( \sum_{i=1}^n a_i^2 \right)^{\frac{3}{2}}}{\sqrt{3\left( \sum_{i=1}^n a_i^2 \right)^{2}}} = \frac{1}{\sqrt{3}} \left( \sum_{i=1}^n a_i^2 \right)^{\frac{1}{2}}.$$

\section{Proof of Lemma \ref{2ndbound}}\label{proof_2ndbound}

Let $b_{t,k} = \sign[V(\pi_{\vtheta_{t,k}'})-V(\pi_{\vtheta_t})]$ and $w_{t,k}:=\langle\nabla_\vtheta V\left(\pi_{\vtheta_t}\right),{\vv}_{t,k}\rangle$. and our objective becomes: 
\begin{align*}
    D_{t,k}:=&\left\langle\nabla_\vtheta V\left(\pi_{\vtheta_t}\right),\left(\hat{{O}}_{t,k}-\operatorname{sign}\left[V\left(\pi_{\vtheta_{t,k}'}\right)-V\left(\pi_{\vtheta_t}\right)\right]\right){\vv}_{t,k}\right\rangle.\\
    =& \left(\hat{{O}}_{t,k}- b_{t,k}\right)w_{t,k}.
\end{align*} 
The randomness of $D_{t,k}$ comes from the perturbation $\vv_{t,k}$ and the panel vote. We first focus on the panel vote with fixed $\vtheta_t$ and $\vtheta_{t,k}'$:
\begin{align}
    \left| \E \left[ \left. D_{t,k} \right| \vtheta_t, \vtheta_{t,k}' \right] \right|
    \leq& \left| \E \left[ \left. \hat{{O}}_{t,k} - b_{t,k} \right| \vtheta_t, \vtheta_{t,k}' \right] \right| \left| \left\langle \nabla_\vtheta V\left(\pi_{\vtheta_t}\right), {\vv}_{t,k} \right \rangle \right|\nonumber\\
    =& 2 \sP \left( \left. \hat{{O}}_{t,k} \neq b_{t,k} \right|  \vtheta_t, \vtheta_{t,k}' \right) |w_{t,k}|
    \label{probstatement}
\end{align}
Let $p(\vtheta_t, \vtheta_{t,k}') = \sP ( \left. \hat{{O}}_{t,k} \neq b_{t,k} \right|  \vtheta_t, \vtheta_{t,k}')$. Thus, the key is to either control the probability $p(\vtheta_t, \vtheta_{t,k}')$ or control the absolute value of inner product $\langle \nabla_\vtheta V\left(\pi_{\vtheta_t}\right), {\vv}_{t,k} \rangle$. According to Algorithm~\ref{oracle}, $\hat{O}_{t,k}$ is the majority vote of preferences over $N$ batches. Let the preference of each batch be $o_{t,k,n}$, with expectation $\E[o_{t,k,n}] = p_{t,k}$. When the perturbed policy has a larger value function, we have:
\begin{align}\label{eq:major-vote-prob-bound}
    p\left(\vtheta_t, \vtheta_{t,k}'\right)
    = &\sP\left(\sum_{n=1}^No_{t,k,n}\leq\frac{N}{2}\right)\nonumber\\
    =&\sP\left(\frac{1}{N}\sum_{n=1}^No_{t,k,n}-\E[o_{t,k,n}] \leq - \left(p_{t,k} - \frac{1}{2} \right)\right).
\end{align}
When $p_{t,k}>1/2$, we can use Hoeffding's inequality to control the probability. However, due to the indistinguishable nature of preferences versus value function sign, this is guaranteed only when the value function difference $| V ( \pi_{\vtheta_{t,k}'} ) - V ( \pi_{\vtheta_t} ) |$ is much larger than $\epsilon^*_D$. This is dependent on how well the sampled perturbation aligns with the gradient direction, so we define such events as follows: 
\begin{align*}
    \gE_{t,k}^+ := &\left\{w_{t,k} \geq \mu_{t}L\left\|{\vv}_{t,k}\right\|_2^2+\frac{\epsilon^*_D}{\mu_{t}} \right\},\\
    \gE_{t,k}^-:= &\left\{ w_{t,k} \leq - \mu_{t}L\left\|{\vv}_{t,k}\right\|_2^2 - \frac{\epsilon^*_D}{\mu_{t}} \right\},\\
    \gE_{t,k}:= &\left\{ \left|w_{t,k}\right|\leq \mu_{t}L\left\|{\vv}_{t,k}\right\|_2^2+\frac{\epsilon^*_D}{\mu_{t}} \right\}.
\end{align*}
On the event $\gE_{t,k}^+$, we indeed have by Lemma~\ref{smoothapproximation} that:
\begin{align*}
    &V\left(\pi_{\vtheta_{t,k}'}\right)-V\left(\pi_{\vtheta_t}\right) \geq  \left\langle\nabla_\vtheta V\left(\pi_{\vtheta_t}\right), \mu_t{\vv}_{t,k}\right\rangle - \frac{\mu_{t}^2L}{2}\left\|{\vv}_{t,k}\right\|_2^2 \\
    =& \mu_t w_{t,k} - \frac{\mu_{t}^2L}{2}\left\|{\vv}_{t,k}\right\|_2^2
    \geq \epsilon^*_D+\frac{\mu_{t}^2L}{2}\left\|{\vv}_{t,k}\right\|_2^2 \geq \epsilon^*_D.
\end{align*}
Then, we can lower-bound the expectation $p_{t,k}$ according to the definition of $\epsilon^*_D$ in definition~\ref{HFlimitation} as follows:
\begin{equation*}
    \begin{aligned}
        p_{t,k} - \frac{1}{2} &=\E_{\gD_1\sim\pi_{\vtheta_{t,k}'},\gD_2\sim\pi_{\vtheta_t}}\left[\varsigma_P\left(\bar{r}\left(\gD_1\right)-\bar{r}\left(\gD_2\right)\right)\right]\\
        &\ge\frac{1}{2}\varsigma_P\left(\frac{1}{2}\left[ V\left(\pi_{\vtheta_{t,k}'}\right)-V\left(\pi_{\vtheta_t}\right) \right]\right).
    \end{aligned}
\end{equation*} 
Therefore, with Hoeffding's inequality, we can upper bound equation~\ref{eq:major-vote-prob-bound} on $\gE_{t,k}^+$ as follows:
\begin{align*}
    p\left(\vtheta_t, \vtheta_{t,k}'\right)
    \leq&\exp\left(-\frac{N}{2}\varsigma_P^2\left(\frac{1}{2} V\left(\pi_{\vtheta_{t,k}'}\right)-V\left(\pi_{\vtheta_t}\right)\right)\right)\nonumber\\
    \leq& \exp\left(-\frac{N}{2}\varsigma_P^2\left(\frac{\mu_{t}w_{t,k}}{4}+\frac{\epsilon_D^*}{4}\right)\right).
\end{align*} 
where the last inequality is because $\varsigma_P(x)$ is monotonically non-decreasing when $x>0$, and on $\gE_{t,k}^+$, we have by Lemma~\ref{smoothapproximation}:
\begin{align*}
    V\left(\pi_{\vtheta_{t,k}'}\right)-V\left(\pi_{\vtheta_t}\right)
    \geq & 2 \cdot \frac{\left\langle\nabla_\vtheta V\left(\pi_{\vtheta_t}\right), \mu_t{\vv}_{t,k}\right\rangle}{2} - \frac{\mu_{t}^2L}{2}\left\|{\vv}_{t,k}\right\|_2^2 \\
    \geq & 2 \frac{\mu_t w_{t,k}}{2} - \frac{\mu_{t}^2L}{2}\left\|{\vv}_{t,k}\right\|_2^2 = \frac{\mu_t w_{t,k}}{2} + \epsilon_D^*.
\end{align*}
Therefore, our target in equation~\ref{probstatement} can be bounded as:
\begin{align*}
    &\left| \E \left[ \left. D_{t,k} \right| \vtheta_t, \vtheta_{t,k}' \right] \right| \leq 2 \exp\left(-\frac{N}{2}\varsigma_P^2\left(\frac{\mu_t|w_{t,k}|}{4}\right)\right)|w_{t,k}|\nonumber\\
    =& 2\exp\left(-\frac{N}{2}\varsigma_P^2\left(\frac{\mu_t|w_{t,k}|}{4}\right)\right)|w_{t,k}|\mathbbm{1}_{\left\{|w_{t,k}|\geq\frac{4}{\mu_t}\varsigma_P^{-1}\left(\sqrt{\frac{4}{N}}\right) \right\}}\nonumber\\
    &+2\exp\left(-\frac{N}{2}\varsigma_P^2\left(\frac{\mu_t|w_{t,k}|}{4}\right)\right)|w_{t,k}|\mathbbm{1}_{\left\{|w_{t,k}|<\frac{4}{\mu_t}\varsigma_P^{-1}\left(\sqrt{\frac{4}{N}}\right) \right\}}\nonumber\\
    \leq&\frac{2}{e^2}|w_{t,k}|+\frac{8}{\mu_t}\varsigma_P^{-1}\left(\sqrt{\frac{4}{N}}\right).
\end{align*}
On event $\gE_{t,k}^-$, we can obtain the same upper bound, but on event $\gE_{t,k}$, we directly bound the probability $p(\vtheta_t, \vtheta_{t,k}')$ trivially by $1$ and results in the following bound:
\begin{align*}
    \left| \E \left[ \left. D_{t,k} \right| \vtheta_t, \vtheta_{t,k}' \right] \right| \leq 2|w_{t,k}| \leq 2\mu_{t}L\left\|{\vv}_{t,k}\right\|_2^2+\frac{2\epsilon^*_D}{\mu_{t}}
\end{align*}
Therefore, in all events, we obtain the following bound:
\begin{align*}
    &\left| \E \left[ \left. D_{t,k} \right| \vtheta_t, \vtheta_{t,k}' \right] \right| \\
    \leq & \frac{2}{e^2}|w_{t,k}|+\frac{8}{\mu_t}\varsigma_P^{-1}\left(\sqrt{\frac{4}{N}}\right) + 2\mu_{t}L\left\|{\vv}_{t,k}\right\|_2^2+\frac{2\epsilon^*_D}{\mu_{t}}.
\end{align*}
Then, we move on to consider the randomness of the perturbation $\vv_{t,k}$ as follows:
\begin{align*}
    &\left|\E\left[\left.D_{t,k}\right|\gF_t\right]\right| \leq  \E\left[ \left| \E \left[ \left. D_{t,k} \right| \vtheta_t, \vtheta_{t,k}' \right] \right| \right] \\
    \leq&\frac{2}{e^2}\E_{{\vv}_{t,k}}\left[|w_{t,k}|\right]+\frac{8}{\mu_t}\varsigma_P^{-1}\left(\sqrt{\frac{4}{N}}\right)+\E_{\vv_{t,k}} \left[ 2\mu_{t}L\left\|{\vv}_{t,k}\right\|_2^2 \right] + \frac{2\epsilon^*_D}{\mu_{t}}\\
    \leq&\frac{2}{e^2}\left\|\nabla_\vtheta V\left(\pi_{\vtheta_t}\right)\circ{e}_{\gI_k}\right\|_2 +2\left(\mu_t L|\gI_k|+\frac{\epsilon^*_D}{\mu_t}\right)+\frac{8}{\mu_t}\varsigma_P^{-1}\left(\sqrt{\frac{4}{N}}\right).
\end{align*} where the last inequality holds by applying Lemma~\ref{Khintchine}.

\end{document}